\documentclass[11pt]{article}

\usepackage[letterpaper, margin=1in]{geometry}
\usepackage[utf8]{inputenc}
\usepackage[T1]{fontenc}
\usepackage{lmodern}
\usepackage[numbers]{natbib}        
\usepackage{amsfonts}
\usepackage{amsmath}
\usepackage{nicefrac}
\usepackage{booktabs}
\usepackage{microtype}
\usepackage{xcolor}
\usepackage{graphicx}
\usepackage{algorithm}
\usepackage{algpseudocode}
\usepackage{authblk}
\usepackage{titlesec}
\usepackage[hypertexnames=false, colorlinks=true, linkcolor=blue,
            citecolor=blue, urlcolor=blue]{hyperref}
\usepackage{url}
\usepackage{orcidlink}

\titleformat{\section}{\normalfont\large\bfseries}{\thesection}{0.6em}{}
\titleformat{\subsection}{\normalfont\bfseries}{\thesubsection}{0.5em}{}
\title{\bfseries Directional Curvature from Armijo Backtracking:\\
A Low-Cost Sharpness Probe and a Calibration-Free\\
Learning-Rate Safeguard for Adam}

\author[1]{Ashmitha R\,\orcidlink{0009-0004-4693-4965}}
\author[2]{J\"org Frochte\,\orcidlink{0000-0002-5908-5649}}

\affil[1]{Department of Artificial Intelligence and Data Science,
Sri Ramakrishna Engineering College,
Anna University,
Coimbatore 641022, Tamil Nadu, India\\
\texttt{ashmitha.2311011@srec.ac.in}}

\affil[2]{Interdisciplinary Institute for Applied AI
and Data Science Ruhr (AKIS),
Department of Electrical Engineering and Computer Science,
Bochum University of Applied Sciences,
Am Hochschulcampus 1,
44801 Bochum, Germany\\
\texttt{joerg.frochte@hs-bochum.de}}

\date{\small Published in \emph{Transactions on Machine Learning Research} (\month/\year):\\
\tmlrpaper}

\usepackage{placeins}   

\def\month{10}  
\def\year{2026}
\def\tmlrpaper{\url{https://openreview.net/pdf?id=foLGwG3ne4}}
\def\coderepo{https://github.com/jfrochte/armijo-curvature-probe}

\begin{document}
\maketitle

\begin{abstract}
Local sharpness, defined by the largest Hessian eigenvalue $\lambda_1$,
sets the maximum stable gradient update size, but its computation would usually
require running Lanczos or Hessian-vector products. However, we notice that even
a single Armijo backtracking line search already contains this
information with just a few forward passes, as the accepted step $\alpha$
determines the \emph{directional} curvature along the search direction up
to the multiplicative band set by the backtracking factor. In other words,
the measured quantity is the average curvature across the tested step,
which empirically turns out to be equal to the pointwise value
$q = g^\top H g/\|g\|^2$ to within the band. The
correlation between $\log\alpha$ and $\log\lambda_1$ on CIFAR-10,
Fashion-MNIST and Imagenette datasets reaches $-0.91$ to $-0.95$ in Pearson
correlation, and even after removing the trend per run the correlation
remains at $-0.60$ to $-0.70$. This allows for a
cheap online Edge-of-Stability estimate of the slow sharpness component. The
employed probing mechanism searches along Adam's first-step update
direction, re-evaluated at the current parameters, at initialisation and
nine times over the course of the first 50 optimiser steps. The learning-rate cap is set as twice the smallest
observed step size, and in the studied learning-rate ranges ($10^{-3}$
to $3.0$) and GPT-2 pretraining experiments all capped runs avoid
divergence; in the
general architecture analysis, one MLP architecture is still sensitive
to the initial batch order. This probing protocol incurs an
approximately one percent overhead, and using a non-binding cap means
that the optimiser's state and first update are bit-identical. There
is no fine-tuning of any of the protocol parameters to any specific
architecture; this is the sense in which this is a
calibration-free safeguard. It is meant as a way of avoiding divergence,
not achieving accuracy. For example, on AG News, a capped run at an
aggressive learning rate still stays at chance-level accuracy. At GPT-2 scale,
multiple measurements also illustrate why an initialisation-only cap
is insufficient: directional curvature rises substantially in the first
five optimiser steps, which motivates the short probationary window.
\end{abstract}

\section{Introduction}

Adam~\citep{kingma2015adam} and its variants
(AdamW~\citep{loshchilov2019adamw}, NAdam~\citep{dozat2016nadam},
AMSGrad~\citep{reddi2018amsgrad}) are the workhorses of contemporary deep
learning. Their per-parameter adaptive scaling absorbs much of the
heterogeneity of neural-network gradients and, in practice, decouples
optimisation from the worst aspects of architecture-specific gradient
geometry. What they do not absorb is the user's choice of \emph{initial}
learning rate. An $\eta$ chosen one order of magnitude too large drives
the loss to numerical infinity within a handful of steps, well before
Adam's running estimates of the second moment have accumulated enough
mass to dampen the update. The result is a single divergent first step,
after which the entire training run is wasted.

The community has developed several pragmatic responses to this problem.
The most principled is the explicit learning-rate range
test~\citep{smith2017cyclical, smith2019superconvergence}, which trains a
model for one or more epochs while sweeping the learning rate and
inspects the loss--learning-rate curve for the largest still-stable
value. A second line of work removes the learning rate from the user
interface entirely by inferring it from running estimates of
online-convex-optimisation quantities;
D-Adaptation~\citep{defazio2023dadaptation},
Prodigy~\citep{mishchenko2023prodigy},
Mechanic~\citep{cutkosky2023mechanic} and the Schedule-Free
family~\citep{defazio2024scheduleFree} are the most recent and most
visible representatives. A third, far less elegant but in practice
ubiquitous approach is for users to manually iterate over a few
candidate values and to discard whatever diverges; we quantify this
retry-on-divergence heuristic as a baseline in
Section~\ref{sec:results-rescue} (Table~\ref{tab:retry}). None of the three
provides a low-cost, calibrate-once safety check that a particular chosen
$\eta$ is safe.

We propose a fourth route, designed to coexist with rather than replace
any of the above. Before the first Adam update, we run a single Armijo
backtracking line search~\citep{armijo1966} at the randomly initialised
parameters: starting from $\alpha = 1$, we backtrack along the negative
initial gradient until the standard sufficient-decrease condition is
satisfied. The resulting step size $\bar\alpha_{\text{init}}$ is a
one-batch empirical estimate of the largest stable gradient step at
initialisation. If the user's $\eta_{\text{init}}$ exceeds $\kappa\cdot
\bar\alpha_{\text{init}}$ for a small safety constant $\kappa$, the
optimiser silently lowers $\eta_{\text{init}}$ to $\kappa\cdot
\bar\alpha_{\text{init}}$ before its first update; otherwise it leaves
the user's choice untouched. The probe costs approximately one fresh
forward-and-backward pass plus a handful of forward-only backtracking
evaluations, totalling under fifty milliseconds for CIFAR-10/ResNet-18
on a single RTX 6000.
The intervention is limited to this initial cap; all subsequent updates
use unmodified Adam. The probe comes in two variants with distinct roles in this
paper. The \emph{raw-gradient} probe admits a clean link to the Hessian
spectrum and is our measuring instrument throughout; the
\emph{direction-matched} probe, which searches along Adam's first-step
update direction, needs no per-architecture calibration and is the variant we
recommend as the default safeguard, with a single fixed safety factor
$\kappa = 2$ across every architecture we test -- selected once on a
nine-architecture suite and confirmed on genuinely held-out ones
(Tables~\ref{tab:kappa-universal-sweep} and~\ref{tab:heldout-kappa2}; at
language-model scale this requires
the probationary window of Section~\ref{sec:method}).

The central claim of this paper is therefore first a \emph{measurement}
claim, and only then a practical one. On a convex quadratic loss, exact
line search along the negative gradient gives $\alpha^\star = 1/q$, where
$q = g^\top H g/\|g\|^2 \le \lambda_1$ is the directional curvature; equivalently
$1/\alpha^\star$ reads curvature along the gradient and is upper-bounded by
$\lambda_1$. We show this textbook relation survives into deep networks in a
precise, usable form
(Section~\ref{sec:method},~\ref{sec:results-mechanism}): the Armijo step
empirically inverts the \emph{directional} curvature $q = g^\top H g/\|g\|^2$
up to the multiplicative backtracking band, and $q$ in turn tracks $\lambda_1$,
so across CIFAR-10/ResNet-18, Fashion-MNIST/CNN and
Imagenette/ResNet-18 (three seeds each,
about 200 step--probe pairs in total) $\log\alpha$ and $\log\lambda_1$ correlate
at Pearson $-0.91$ to $-0.95$, and the correlation survives a per-run
trend-robustness check at $-0.60$ to $-0.70$
(Table~\ref{tab:correlation-detrended}). The line search is thus a low-cost online
sharpness sensor ($\sim 5$ forward passes, no Hessian-vector products),
giving an Edge-of-Stability reading~\citep{cohen2021eos,
lewkowycz2020catapult, damian2023eos, andreyev2024eoss} that the
literature normally obtains with Lanczos iteration.

We then ask what this measurement buys in practice.
Used once at initialisation, the probe caps a too-large learning rate:
on CIFAR-10/ResNet-18 vanilla Adam diverges in the first step for any
$\eta \ge 0.1$, whereas the init-probe variant holds test accuracy in
$[0.79, 0.83]$ across $\eta \in [10^{-3}, 3.0]$ (5 seeds) at about one
percent overhead. Two more elaborate
periodic controllers (\emph{watchdog}, \emph{tracker}) match the one-shot
probe on CIFAR-10 but do not transfer to other settings and add no robust
benefit; we therefore recommend the one-shot probe.

In summary, this paper contributes: \emph{(i)} the observation, with an
exact bracketing of the segment-averaged directional curvature (of $q$
itself on a quadratic model) and a strong empirical
$\alpha$--$\lambda_1$
correlation across three architectures, that the Armijo line-search step is
a low-cost, Hessian-free online estimate of local curvature, in effect
an Edge-of-Stability reading for the price of a few forward passes
(Section~\ref{sec:results-mechanism}); \emph{(ii)} a one-shot \emph{init
probe} built on this measurement that caps Adam's initial learning rate, a
low-overhead, calibrate-once-then-deploy safeguard against learning-rate
misspecification (Section~\ref{sec:method});
and \emph{(iii)} a controlled
delineation of where learning-rate control helps and where it does not,
across CIFAR-10, Fashion-MNIST, Imagenette, AG News and GPT-2
pretraining on WikiText-103, and against gradient
clipping~\citep{pascanu2013gradient},
warmup~\citep{goyal2017largeminibatch} and a retry-on-divergence
baseline.

We make two claims. The first is a measurement claim: the Armijo step
reads local curvature. Wherever the backtracking constraint binds,
$1/\alpha$ brackets the directional curvature $q$ within the
factor-$1/\beta$ band, and it tracks the top eigenvalue $\lambda_1$ both
as a static correlation (Table~\ref{tab:correlation}) and online over
training (Figure~\ref{fig:tracking}), at the fidelity characterised in
Table~\ref{tab:fidelity}; this $\alpha$--$\lambda_1$ link is a
within-run effect on the tested architectures, not a universal law. The
second is a divergence-prevention claim: used as a one-shot cap, the measurement makes
Adam robust to a too-large initial learning rate where that rate is
genuinely unsafe (Tables~\ref{tab:rescue}
and~\ref{tab:default-unsafe}), it is a no-op otherwise, and a
probationary variant of the same measurement extends the divergence
prevention to GPT-2 pretraining (Table~\ref{tab:lm-rescue}). Both claims
concern training stability rather than optimisation speed; periodic
in-training probing, in particular, adds no robust benefit and can hurt
(Table~\ref{tab:periodic}).

\section{Related work}

The oldest pragmatic response to learning-rate sensitivity is the range
test. \citet{smith2017cyclical, smith2019superconvergence} introduced an
explicit range test in which a model is trained for a few epochs while
the learning rate is exponentially increased; the inflection point of the
loss curve indicates the largest safe learning rate. Range tests are
robust and widely adopted, but they require one or more epochs of
training as upfront cost. Our probe is structurally similar in spirit but
strictly lighter: a single backtracking line search at initialisation,
consisting of one forward--backward pass to obtain the direction plus on
average five forward-only loss evaluations for backtracking.

A more radical recent line of work removes the learning rate from the
interface altogether, proposing optimisers that automatically infer their
own learning rate from running estimates of online-convex-optimisation
quantities. D-Adaptation~\citep{defazio2023dadaptation} maintains a
running lower bound on the distance to a minimiser and uses it as the
learning rate; the method is hyperparameter-free and provably matches the
optimal convergence rate for convex Lipschitz problems.
Prodigy~\citep{mishchenko2023prodigy} extends D-Adaptation by improving
its convergence rate by a $\sqrt{\log(D/d_0)}$ factor.
Mechanic~\citep{cutkosky2023mechanic} tunes the scale factor of \emph{any}
base optimiser online, again grounded in online-convex-optimisation
reductions. The Schedule-Free family~\citep{defazio2024scheduleFree}
removes both the learning-rate \emph{schedule} and the requirement to
specify the stopping time. All four methods modify the optimiser's
internal state continuously; in contrast our probe sits next to Adam,
runs once at initialisation (optionally every $K_{\text{probe}}$
steps), and only adjusts a single scalar (Adam's $\eta$).

A third family assumes the base rate is already known and only shapes it
over time. Cyclical learning rates~\citep{smith2017cyclical}, SGDR~\citep{loshchilov2017sgdr}
and super-convergence~\citep{smith2019superconvergence} prescribe schedules
on top of a known base learning rate. Linear warmup~\citep{goyal2017largeminibatch}
is the simplest example and is the closest baseline to our init probe;
it is reactive (gradients can still explode if the warmup target is
too high) rather than calibrated.

Closest to our mechanism is the stochastic line-search literature.
\citet{vaswani2019sls} (SLS) apply Armijo backtracking on
each mini-batch. SLS uses the line search for the actual update, while we
use it only as a probe and let Adam do the actual optimisation. The Polyak
step of \citet{loizou2021sps} and the parabolic
approximation~\citep{mutschler2020pal} are alternative non-standard
adaptive step rules.

The reason a line-search probe carries so much information about the
stable range is best understood through the Edge-of-Stability literature.
\citet{cohen2021eos} showed that \emph{full-batch} gradient
descent on a deep network at fixed $\eta$ enters a regime where the top
Hessian eigenvalue $\lambda_1$ rises until $\lambda_1 \approx 2/\eta$ and
then oscillates around that value. The catapult
mechanism~\citep{lewkowycz2020catapult}, the theoretical analysis of
\citet{arora2022eos}, and the implicit-bias study
of \citet{damian2023eos} are part of the same line. Most relevant to us
is the recent extension to mini-batch SGD by \citet{andreyev2024eoss}, who show that what stabilises at
$2/\eta$ in stochastic optimisation is not $\lambda_1$ but
\emph{Batch Sharpness}, the expected directional curvature of the
mini-batch Hessian along the corresponding stochastic gradient. Our
probe-batch line search measures essentially this directional curvature,
which is why the empirical correlation we observe (Pearson $-0.91$ to
$-0.95$ on log scale) is so tight in the mini-batch regime. The empirical
methodology of all of these works relies on
PyHessian-style~\citep{yao2020pyhessian} Lanczos / power iteration for
explicit eigenvalue estimation, which is computationally expensive.

A closely related recent line reads curvature from forward passes rather
than from Hessian-eigenvalue iterations. \citet{kalra2024warmup} choose
$\eta_{\text{init}}$ via the loss catapult mechanism, addressing the same
too-large-initial-rate failure as our safeguard but by a different
mechanism, and \citet{kalra2026critical} design a two-phase line search
along the update direction to estimate a \emph{critical sharpness}
$\lambda_c = 2/\eta_c$, which they use as a training-time diagnostic that
tracks progressive sharpening and the Edge of Stability at LLM scale. Our
finding is complementary. Rather than a purpose-built measure, the
curvature signal is already carried by a standard Armijo
sufficient-decrease backtracking search: the accepted step brackets the
directional curvature within the multiplicative band of the backtracking
factor, so a practitioner already running such a line search obtains the
reading at no extra cost. We further turn a single reading at
initialisation into a calibration-free learning-rate safeguard
($\kappa = 2$; calibration-free in the sense defined in the abstract, that
no constant of the protocol is tuned per architecture) rather than a
diagnostic. Section~\ref{sec:results-lm}
connects the two views directly: at GPT-2 initialisation, repeated
probes show that the accepted step decreases sharply within the first
five optimiser steps as curvature increases, which is what forces the
probationary variant of the safeguard there.

Building on this curvature view, the closest related work is \citet{roulet2024stepping},
who analyse classical line-search and quadratic-fit tuners (PAL,
Armijo backtracking) under the lens of curvature dynamics. They find
empirically that classical tuners ``undershoot the edge of
stability'', producing a snowball effect of ever-increasing sharpness
and ever-decreasing learning rates; this matches what we observe for
GS+Armijo on CIFAR-10/ResNet-18 (Appendix~\ref{app:gs-failure}). They
propose CDAT, a continuous tuner that dynamically drives $\eta$ towards
the EoS. Our work is complementary: we focus on the mini-batch regime,
treat the line search as a probe rather than as a tuner, and design our
controllers (Watchdog, Tracker) to sit deliberately \emph{below} the
EoS at $\kappa\,\bar\alpha \approx \bar\alpha/4$, which trades a small
amount of optimisation efficiency for divergence safety. Our practical
contribution, the init-probe divergence-prevention mechanism, is to our knowledge not
present in their work.

Finally, the simplest and most widely deployed defence is gradient
clipping: \citet{pascanu2013gradient} introduced global gradient-norm
clipping to combat exploding gradients in RNN training. Clipping bounds the
update norm but not the implicit step into a sharp region, and we compare
against it directly in our experiments.

\section{Method}
\label{sec:method}

Throughout this section we denote the model parameters by $\theta$, the
loss on a mini-batch $\mathcal{B}$ by $L(\theta;\mathcal{B})$, and Adam's
scalar learning rate by $\eta$. We assume the standard Adam update
$\theta_{t+1} = \theta_t - \eta\, m_t/(\sqrt{v_t}+\varepsilon)$ with the
default first- and second-moment estimates~\citep{kingma2015adam}; our
controller never modifies these internal estimates and only adjusts the
scalar $\eta$.

The core of our method is a single Armijo backtracking line
search~\citep{armijo1966} executed once, immediately before the first
Adam update. Concretely, we sample the first training mini-batch
$\mathcal{B}_0$, compute the loss $L_0 = L(\theta_0;\mathcal{B}_0)$ and
gradient $g_0 = \nabla_\theta L(\theta_0;\mathcal{B}_0)$ at the random
initialisation $\theta_0$, and search for the largest step
$\alpha\in(0,1]$ along the descent direction $d_0 = -g_0$ that
satisfies the standard sufficient-decrease (Armijo) condition
$L(\theta_0 + \alpha d_0;\mathcal{B}_0) \le L_0 - c\,\alpha\,\|g_0\|^2$.
We initialise $\alpha = 1$ and shrink by a factor $\beta = \tfrac12$
until the condition holds, capping the iteration at $K = 8$ backtracks.
The resulting $\bar\alpha_{\text{init}}$ is a single-batch empirical
estimate of the largest stable gradient step at initialisation.
Algorithm~\ref{alg:init-probe} states the procedure compactly.

\begin{algorithm}[ht]
\caption{Init probe with safety cap (general descent direction $d$).}
\label{alg:init-probe}
\begin{algorithmic}[1]
\State \textbf{Input:} model $\theta_0$, loss $L$, mini-batch $\mathcal{B}_0$,
       direction rule $d$, safety factor $\kappa$, Armijo constants $c$, $\beta$,
       maximum backtracks $K$ (deployed constants per context:
       Table~\ref{tab:protocol}).
\State \textbf{Variants:} raw-gradient $d_0 = -g_0$ (mechanism, $\kappa = 0.25$);
       \textbf{direction-matched (recommended)} $d_0 = -g_0/(|g_0| + \varepsilon)$
       (Adam's first-step geometry, $\kappa = 2$).
\State $L_0 \gets L(\theta_0; \mathcal{B}_0)$, $g_0 \gets \nabla L(\theta_0; \mathcal{B}_0)$
\State $\alpha \gets 1$
\For{$k = 1, \dots, K$}
  \If{$L(\theta_0 + \alpha d_0; \mathcal{B}_0) \le L_0 + c\,\alpha\, g_0^\top d_0$}
    \State \textbf{break} \Comment{Armijo holds: accept the tested $\alpha$}
  \Else
    \State $\alpha \gets \beta \alpha$ \Comment{shrink; if $k = K$ this last $\alpha = \beta^{K}$ is returned untested}
  \EndIf
\EndFor
\State \textbf{on saturation} (no candidate accepted): either keep the
       untested $\alpha = \beta^{K}$ and cap conservatively at
       $\kappa\beta^{K}$, or \emph{extend} the backtracking until a
       candidate is accepted \Comment{per-context choice, Table~\ref{tab:protocol}; see the band derivation}
\State $\bar{\alpha}_{\text{init}} \gets \alpha$
\State \textbf{if} $\eta_{\text{user}} > \kappa\,\bar{\alpha}_{\text{init}}$
       \textbf{then} $\eta \gets \kappa\,\bar{\alpha}_{\text{init}}$ \textbf{else} $\eta \gets \eta_{\text{user}}$
\State Restore $\theta_0$; start Adam at $\eta$.
\end{algorithmic}
\end{algorithm}

For the raw-gradient probe on CIFAR-10/ResNet-18 we use
$\kappa = 0.25$. We arrived at this number directly from the empirical
divergence boundary of vanilla Adam reported in
Appendix~\ref{sec:results-divergence}: vanilla Adam first diverges at
$\eta \approx 0.6\,\bar\alpha_{\text{warmup}}$, the warmup-averaged anchor
measured there, so a safety factor of
$\kappa = 0.25$ leaves a comfortable margin of roughly $2\times$
without being overly conservative. The choice is forgiving: a sweep over
$\kappa$ (Appendix~\ref{app:kappa-sensitivity}) shows that any
$\kappa \le 0.25$ preserves divergence prevention on CIFAR-10, with graceful rather
than catastrophic degradation beyond. \emph{For the raw-gradient probe}, as
Section~\ref{sec:results-fashion} will show, $\kappa$ is architecture-specific
and a $\sim$one-minute divergence sweep is the recommended way to set it on a
new architecture (once chosen it can be reused across hyperparameter searches
and seeds); the direction-matched variant introduced next removes this
dependence and uses a single fixed $\kappa = 2$ throughout.

\textbf{Direction-matched Adam probe.} The probe so far line-searches along
$d_0 = -g_0$, but Adam does not step along the raw gradient. Its first update
direction is $-m_0/(\sqrt{v_0}+\varepsilon)$; with bias correction at $t=1$
this is $\hat m_0 = g_0$ and $\hat v_0 = g_0^2$, so Adam's first step points
along
\[
  d_{\text{Adam}} \;=\; -\,\frac{g_0}{\sqrt{g_0^2}+\varepsilon}
  \;=\; -\,\frac{g_0}{|g_0|+\varepsilon}
  \;\approx\; -\operatorname{sign}(g_0),
\]
the preconditioned direction in which the optimiser actually moves. The
direction-matched variant of the probe runs the \emph{same} backtracking
search along $d_{\text{Adam}}$ instead of $-g_0$, accepting the largest
$\alpha\in(0,1]$ that satisfies the general Armijo condition
$L(\theta_0 + \alpha\, d_{\text{Adam}};\mathcal{B}_0)
 \le L_0 + c\,\alpha\, g_0^\top d_{\text{Adam}}$
(note $g_0^\top d_{\text{Adam}} < 0$), and caps the learning rate at
$\eta \le \kappa\,\bar\alpha_{\text{Adam}}$ exactly as in
Algorithm~\ref{alg:init-probe}. Because $\bar\alpha_{\text{Adam}}$ is now
measured in Adam's own geometry, the conversion factor $\kappa$ no longer has
to absorb the per-architecture gradient scale: a single fixed value transfers
across architectures, which we establish empirically in
Section~\ref{sec:results-fashion} (Table~\ref{tab:universal-kappa}). This is
the variant we recommend when a per-architecture divergence sweep is
undesirable; the raw-gradient probe is retained in the body experiments
because it admits the clean link to $\lambda_1$ used to expose the mechanism.

Two protocol details matter beyond the classifier benchmarks and cost
little to state generally. First, when the probe is used for capping
and \emph{saturates}, that is, no candidate is accepted within the $K$
backtracks, deployment extends the backtracking until a step is
accepted: saturation itself is the signal, no threshold is involved,
and only an accepted step certifies a cap. The no-normalisation ResNet
of Table~\ref{tab:default-unsafe} is the architecture that needs this;
its stable rate lies below $\kappa\beta^{K}$, and with the extended
reading the same $\kappa = 2$ prevents divergence. (The divergence-prevention tables of
Section~\ref{sec:results-rescue} use the plain $K = 8$ probe that on
saturation returns the next, untested candidate $\beta^{K}$, a cap at
least as permissive as the extended reading,
so their zero-divergence results hold a fortiori for the extended
rule, and Table~\ref{tab:kappa-universal-sweep} shows the stricter cap
costs no accuracy.) Second, the cap as described freezes a reading
taken at $\theta_0$. Where the landscape sharpens within the first few
optimiser steps, that reading is stale by the time it is applied
(Section~\ref{sec:results-lm}); we therefore extend the init probe into
a short \emph{probationary window}: the same probe is re-run at steps
$\{1, 2, 3, 5, 8, 12, 20, 35, 50\}$, at the standard depth $K$, applying
the same first-step direction rule $-g_t/(|g_t|+\varepsilon)$ to the
gradient at the current parameters (the probe deliberately does not read
the optimiser's moment state), and the
cap is tightened monotonically to $\kappa \cdot \min_t \bar\alpha_t$,
after which the optimiser runs untouched. The window adds at most nine
probes (under $1\%$ of a $10$k-step run) and can only lower the cap.
The re-probes deliberately keep the floor $\beta^{K}$: deeper
in-training readings track the instantaneous mini-batch step, which on
GPT-2 lies a factor ${\sim}50$ below the empirically safe rate, while
the window's job is only to correct a stale initialisation reading. On
the classifier suite the initialisation reading already sits at or
below the floor (Table~\ref{tab:probe-variance}), so the probationary
window rarely tightens the cap: re-running the full protocol through the nine-architecture
suite of Table~\ref{tab:universal-kappa} gives two divergences in $54$
runs, both the plain MLP at the $100\times$ rate, with accuracies
matching the static cap on the remaining architectures; ten-epoch
spot checks on CIFAR-10 and Fashion-MNIST reproduce the
direction-matched accuracies unchanged. The MLP is the suite's one
marginal architecture at this cap: the \emph{first} optimiser step
under $\kappa\bar\alpha \approx 0.016$ survives or diverges with the
training batch order (a dedicated $26$-run check finds $3/13$ static
and $2/13$ probationary failures), which no init-time window can
prevent; the conservative $\kappa = 1$ halves the cap and shows no
failure in the sweep.

The init probe handles the most common failure mode, a single
catastrophically wrong initial learning rate, but, probationary window
aside, cannot react to sharpness changes later in training. For users who want continuous
in-training adaptation, we provide two periodic controllers that re-run
the same Armijo probe every $K_{\text{probe}}$ Adam steps (default
$K_{\text{probe}} = 50$) and adjust $\eta$ from the resulting
$\alpha_t$. The defensive \emph{Watchdog} maintains a baseline
$\bar\alpha_{\text{base}}$ from the first probe and shrinks $\eta$ by a
factor of two whenever $\alpha_t / \bar\alpha_{\text{base}}$ falls below
$0.5$; symmetrically it grows $\eta$ by a factor of $1.2$ when the
ratio recovers above $0.8$, but never beyond the user's initial value
or below $0.05\,\eta_{\text{init}}$. The active \emph{Tracker} updates
$\eta$ towards the probed step using an exponential moving average,
$\eta_{t+1} = (1-\rho)\eta_t + \rho\,\kappa\alpha_t$, with the same
$\kappa$ as the
init probe and a smoothing weight $\rho = 0.1$ (distinct from the
backtracking factor $\beta$). Both controllers were tuned by inspection
on CIFAR-10/ResNet-18; we will see in
Section~\ref{sec:results-fashion} that their tuned thresholds do not
transfer to a small CNN, and that on Fashion-MNIST it is in fact best
to disable them and rely on the init probe alone.

As for compute, an init probe consumes one extra
forward-and-backward pass at $\theta_0$ (to obtain the descent
direction) plus on average five \emph{forward-only} backtracking
evaluations, a pass count that is architecture-independent; on
CIFAR-10/ResNet-18 the probe totals $\sim 50$ ms on a single RTX 6000.
Because
Adam-InitOnly fires this probe only once, its wall-clock overhead is
about $1\%$ on CIFAR-10/ResNet-18, within measurement noise of
vanilla Adam. A periodic probe is structurally identical; with the
default $K_{\text{probe}}=50$ the periodic-probe controllers add a
measured $\approx\!2.7\%$ (Adam-Tracker) to $\approx\!5.3\%$
(Adam-Watchdog) of wall-clock (Section~\ref{sec:results-cost}).
Evaluating the backtracking trials forward-only, rather than
recomputing a gradient at every trial, roughly halves this periodic
overhead.

\section{Experimental setup}
\label{sec:setup}

Our primary testbed is ResNet-18~\citep{he2016resnet} with the standard
CIFAR adaptation
($3\times 3$ stride-1 conv stem, no max-pool, no pre-trained weights),
trained on CIFAR-10~\citep{krizhevsky2009cifar} (50k train / 10k test) with standard augmentation
(random crop with 4-pixel padding, horizontal flip), batch size 256, for
10 epochs. On this testbed we benchmark five optimisers, all built on top of the same vanilla Adam
implementation (\texttt{torch.optim.Adam} with $\beta_1=0.9$,
$\beta_2=0.999$, $\varepsilon=10^{-8}$). Three are our own contributions
of varying complexity: \emph{Adam-InitOnly}, which performs only the
init probe of Section~\ref{sec:method} and then leaves Adam alone;
\emph{Adam-Watchdog}, which additionally re-probes every fifty Adam
steps and applies the defensive shrink/grow rule; and
\emph{Adam-Tracker}, which uses the same periodic probe but feeds it
into the active EMA controller. Two are simple and widely deployed
baselines that target similar failure modes: \emph{Adam + grad-clip},
which clips the global gradient norm to one before each Adam update,
and \emph{Adam + linear warmup}, which initialises Adam at
$\eta_{\text{init}}/100$ and ramps the learning rate linearly to
$\eta_{\text{init}}$ over the first 200 steps. Two recent parameter-free
optimisers, Schedule-Free AdamW~\citep{defazio2024scheduleFree} and
Prodigy~\citep{mishchenko2023prodigy}, complete the comparison in the
divergence-prevention experiments. Adam-InitOnly rows come in the two probe variants
of Section~\ref{sec:method}: rows marked ``dir.''\ use the recommended
direction-matched probe of Algorithm~\ref{alg:init-probe} at the fixed
$\kappa = 2$; unmarked rows use the raw-gradient probe with its
per-architecture $\kappa$, retained for its clean link to $\lambda_1$.

Transfer beyond this primary setting is tested on three further
benchmarks. Fashion-MNIST~\citep{xiao2017fashionmnist} (60k train / 10k
test) is trained with a
small CNN (two convolutional and two fully-connected layers, 0.4M
parameters); AG News topic classification~\citep{zhang2015charcnn}
(4 classes, 120k/7.6k
train/test) with a small from-scratch encoder-only
Transformer~\citep{vaswani2017transformer} (4
layers, $d_{\text{model}} = 128$, word-level tokeniser); and
Imagenette~\citep{howard2019imagenette}, the fastai ImageNet subset
(10 classes, 9.5k/3.9k train/val images), trained at $128 \times 128$
with the standard full-stem ResNet-18 as a real-resolution control
(Appendix~\ref{app:imagenette}). Probe and
controllers are identical across all four testbeds; only the data
loader and model differ. The Fashion-MNIST CNN also serves as the
second architecture in the mechanism experiments below, and the
cross-architecture study of the direction-matched probe widens the set
to nine models (normalisation variants, an MLP, an LSTM, and
Transformers of several scales), introduced where they appear.

Each (optimiser, initial learning rate) combination of the primary
benchmark grids is run with 5 seeds; the other studies use the seed counts
stated in their captions and in Table~\ref{tab:protocol} (three for the
cross-architecture and correlation studies, two at language-model scale).
We report
mean $\pm$ standard deviation and divergence rate. In the optimiser-comparison
tables, bold marks the best mean per column within each block of rows, since
the blocks compare different regimes (standard baselines, probe controllers,
parameter-free optimisers; uncalibrated versus calibrated); divergent,
partially divergent and chance-level entries are not marked. A run is marked
\emph{diverged} if loss exceeds $5\times$ initial or becomes \texttt{NaN}.
The mini-batch loss is logged \emph{before} each update, so the blow-up
caused by the destabilising first update is recorded at the next logged
step; throughout, ``diverges in the first step'' refers to this first
update, which our logs flag within the first two logged steps.
Hardware: single NVIDIA RTX 6000 Ada Generation per run. The experimental
protocol (seeds, averaging, and how the overhead and curvature numbers
were obtained) is detailed in Appendix~\ref{app:protocol}.

The five primary datasets are public and used in their standard published
form,
with no filtering or re-splitting: CIFAR-10 from the reference
distribution, Fashion-MNIST through \texttt{torchvision}, AG News from the
HuggingFace dataset \texttt{fancyzhx/ag\_news}, Imagenette in the 160px
variant \texttt{imagenette2-160} from the fast.ai mirror, and WikiText-103
in the raw configuration \texttt{wikitext-103-raw-v1} of the HuggingFace
dataset \texttt{Salesforce/wikitext}, tokenised once with the GPT-2 byte-pair
encoding (\texttt{tiktoken} encoding \texttt{gpt2}, vocabulary $50257$).
The released code sets up every dataset from these sources with a single
command, so the exact token stream behind the perplexities reported below is
reproducible.

Because the probe appears in five experimental contexts with different
budgets, Table~\ref{tab:protocol} collects every deployed constant in one
place; the saturation column is discussed with the band derivation in
Section~\ref{sec:results-rescue}. The Armijo constants are global:
$c = 10^{-4}$, $\beta = 1/2$, $\alpha_{\max} = 1$. In all contexts the
probe restores parameters, non-parameter buffers (BatchNorm statistics)
and RNG state after probing, so a non-binding cap leaves the optimiser
state and the first update bit-identical (Appendix~\ref{app:noop}).

\begin{table}[htbp]
\centering
\caption{Deployed probe protocol per experimental context. ``$K=8$,
keep'' returns the next untested candidate $\beta^{8}$ on saturation and
caps at $\kappa\beta^{8}$; ``extend'' continues halving until an
Armijo-acceptable candidate is found (up to depth $30$; every suite run
accepted). Budgets differ deliberately across contexts -- accuracies are
not comparable across tables with different budgets.}
\label{tab:protocol}
\resizebox{\textwidth}{!}{%
\begin{tabular}{llllll}
\toprule
Context (tables) & Direction & Init depth / saturation & $\kappa$ & Re-probes & Budget $\times$ seeds \\
\midrule
Measurement only (\ref{tab:correlation}, \ref{tab:fidelity}; probe in eval mode) & raw-grad & $K=10$ & --- (no cap) & every $K_{\text{probe}}$ steps & 10--20 epochs $\times$ 3 \\
Benchmark divergence-prevention grids (\ref{tab:rescue}, \ref{tab:fashion-rescue}, \ref{tab:agnews-rescue}) & raw-grad / Adam-dir & $K=8$, keep & per-arch / $2$ & none & 10 epochs $\times$ 5 \\
Cross-architecture suite (\ref{tab:universal-kappa}, \ref{tab:kappa-universal-sweep}, \ref{tab:kappa-perarch}) & Adam-dir (+ raw-grad) & extend ($\le 30$) & $3$, sweep $\{1,2,3\}$ / $0.25$ & none & 3 epochs\textsuperscript{a} $\times$ 3 \\
Probation revalidation (Section~\ref{sec:method}) & Adam-dir & extend ($\le 30$) & $2$ & $K=8$ at steps $\{1,\dots,50\}$ & 3 epochs\textsuperscript{a} $\times$ 3 \\
LM pretraining (\ref{tab:lm-rescue}) & Adam-dir & $K=8$, keep & $2$ & $K=8$ at steps $\{1,\dots,50\}$ & $10^{4}$ steps $\times$ 2 \\
\bottomrule
\multicolumn{6}{l}{\footnotesize \textsuperscript{a}600 optimiser steps on AG News.}
\end{tabular}}
\end{table}

\section{Results}
\label{sec:results-rescue}

\phantomsection\label{sec:results-mechanism}
We open with the measurement at the heart of the method: what the
line-search step reveals about the local loss landscape, and at what cost.
The init probe and the periodic probes both estimate, at very low cost,
the largest gradient step that the local loss landscape can absorb. We
make this connection precise, and we do it for a general descent
direction $d$ (with $g^\top d < 0$), because the direction along which the
\emph{optimiser} steps will matter below. On a quadratic loss
$L(\theta) = \tfrac12 (\theta - \theta^\star)^\top H (\theta - \theta^\star)$
with $H \succ 0$, exact line search along $d$ chooses the minimiser
\[
  \alpha^\star \;=\; \frac{-\,g^\top d}{d^\top H d} \;=\; \frac{1}{q_d},
  \qquad
  q_d \;:=\; \frac{d^\top H d}{-\,g^\top d},
\]
where $q_d$ is the curvature \emph{along} $d$. For the raw gradient
$d = -g$ this is $q_{-g} = g^\top H g / \|g\|^2 \in [\lambda_n, \lambda_1]$,
so $1/\lambda_1 \le \alpha^\star \le 1/\lambda_n$. Empirically we measure $\alpha$ and the top Hessian
eigenvalue $\lambda_1$ throughout training of ResNet-18 on CIFAR-10 and,
as cross-architecture checks, of a plain CNN on Fashion-MNIST and of the
standard full-stem ResNet-18 on Imagenette at $128$px. The
results are reported in Table~\ref{tab:correlation}.

\begin{table}[htbp]
\centering
\caption{Empirical correlation between the line-search step $\alpha$ and
the top Hessian eigenvalue $\lambda_1$ along training, on
CIFAR-10/ResNet-18, Fashion-MNIST/CNN and Imagenette with the standard
full-stem ResNet-18 (10 epochs, $K_{\text{probe}}=100$; Imagenette: 20
epochs, $K_{\text{probe}}=30$, since its epochs are only 37 steps).
Means over $3$ seeds where indicated; $p$-value for the log--log Pearson.}
\label{tab:correlation}
\resizebox{\textwidth}{!}{%
\begin{tabular}{lcccr}
\toprule
Variant & $\bar{r}_{\log\alpha,\log\lambda_1}$ & $\bar{r}_{\text{Spearman}}$ & $\bar{p}$ & $\bar\alpha$ \\
\midrule
Pure Armijo, CIFAR-10/ResNet-18 (3 seeds) & $-0.913 \pm 0.013$ & $-0.83 \pm 0.04$ & $< 10^{-7}$ & $0.666$ \\
Pure Armijo, Fashion-MNIST/CNN (3 seeds) & $-0.950 \pm 0.017$ & $-0.89 \pm 0.04$ & $< 10^{-10}$ & $0.292$ \\
Pure Armijo, Imagenette/ResNet-18 (3 seeds) & $-0.907 \pm 0.021$ & $-0.85 \pm 0.03$ & $< 10^{-8}$ & $0.274$ \\
Pure Armijo, Digits/MLP (1 seed) & $-0.775$ & $-0.61$ & $0.024$ & $0.95$ \\
GS+Armijo, CIFAR-10/ResNet-18 (1 seed, Appendix) & $-0.32$ & $-0.17$ & $0.18$ (n.s.) & $0.05$ \\
\bottomrule
\end{tabular}}
\end{table}

The empirical correlation is strong, reproducible across seeds, and highly
significant. Crucially it holds beyond the primary testbed: on
Fashion-MNIST with a plain CNN (no residual
connections) the log--log Pearson coefficient is $-0.950 \pm 0.017$, an
equally strong relationship to the $-0.913 \pm 0.013$ measured on
CIFAR-10/ResNet-18, and on Imagenette at real resolution it is
$-0.907 \pm 0.021$ with the standard full-stem ResNet-18, so the
relation is neither an artefact of one network family nor of
thumbnail-resolution inputs. This is the textbook quadratic relation
$\alpha \approx 1/\lambda$ surviving into the deep-net regime in the
sense made precise by Edge of Stability~\citep{cohen2021eos}: the line
search is, in effect, a continuous online estimator of the same
$\lambda_1$ that the EoS literature measures via expensive Lanczos
iteration~\citep{yao2020pyhessian}.

Both series drift over training. The line search steers the iterate into
progressively flatter regions, so $\log\alpha$ rises while
$\log\lambda_1$ falls, and part of the raw correlation could reflect
this shared trend rather than step-level co-movement.
Table~\ref{tab:correlation-detrended} therefore repeats the analysis
per run with the trend removed, in two ways. Removing a linear-in-step
trend from each log series and correlating the residuals, a partial
correlation given time, gives $-0.60$ to $-0.70$ on the three
architectures; the per-run $p$-values, combined across seeds with
Fisher's method, remain below $10^{-5}$ everywhere. The stricter
first-difference test, which correlates probe-to-probe changes and is
immune to any shared monotone drift, keeps the negative sign on average
but is weak ($-0.26$ to $-0.37$) and reaches the $5\%$ level only on
Imagenette. The weakness has a mechanical reason: backtracking quantises
$\alpha$ to the grid $\{1, \beta, \beta^2, \dots\}$, so changes of
$\lambda_1$ smaller than the factor $1/\beta = 2$ between consecutive
probes frequently leave $\alpha$ unchanged, and differencing amplifies
exactly this quantisation noise. The honest reading of
Tables~\ref{tab:correlation} and~\ref{tab:correlation-detrended}
together is that the probe reliably tracks sharpness across training
including its slow component, which is the regime a learning-rate
safeguard and an Edge-of-Stability monitor operate in, while
fluctuations below the backtracking resolution are not resolved step to
step.

Two objections remain against any statistic computed on these runs: the
trajectories themselves are generated by the line search whose reading
is being validated, and the tests above treat temporally dependent probe
checkpoints as independent (probes at the search ceiling $\alpha =
\alpha_{\max}$ additionally carry no curvature information -- on the
CIFAR-10 Armijo runs $56\%$ of probe points sit at that ceiling).
Table~\ref{tab:correlation-robust} therefore repeats the analysis under
a stricter protocol -- ceiling-censored points excluded,
autocorrelation-robust $p$-values via the effective sample size of
\citet{pyper1998}, AR(1) form -- and, decisively, extends it to
trajectories the line search did \emph{not} generate: fixed-step Adam
at the default and at the largest non-diverging grid rate (the
near-EoS regime), AdamW, and SGD with momentum, with the probe running
as a pure observer (eval mode, frozen probe batch,
Table~\ref{tab:protocol}). The relation survives both corrections. On
the Armijo trajectories the detrended correlations stay negative and
significant under the robust test, though CIFAR-10, with only
${\sim}8$ uncensored points per run, drops to $p = 0.017$. On the
independently generated trajectories, where no ceiling censoring occurs
and ${\sim}40$--$47$ observations per run remain, every cell is
significant in the detrended \emph{and} the first-difference test --
and the near-EoS cells are the strongest of all
(CIFAR-10: detrended $-0.73 \pm 0.07$, $p < 10^{-18}$; first
differences $-0.65 \pm 0.09$, $p < 10^{-12}$), exactly where step and
sharpness should couple most tightly. The correlation is therefore not
an artefact of Armijo-generated trajectories, and the step-level
co-movement that the quantised Armijo steps cannot resolve
(Table~\ref{tab:correlation-detrended}) is resolved by the probe on
optimiser trajectories near the Edge of Stability.

\begin{table}[htbp]
\centering
\caption{Correlation under the stricter protocol: probe points at the
search ceiling excluded (fraction reported), $p$-values
autocorrelation-robust (Fisher-combined across 3 seeds). Top block: the
pure-Armijo mechanism runs of Table~\ref{tab:correlation-detrended}
re-analysed. Bottom block: independently generated fixed-step
trajectories (10 epochs, probe every 50 steps, measurement-only probe).}
\label{tab:correlation-robust}
\resizebox{\textwidth}{!}{%
\begin{tabular}{lccccccc}
\toprule
Trajectory & $n$/seed & cens. & raw $\bar r$ & detrended $\bar r$ & $p_F$ & first-diff $\bar r$ & $p_F$ \\
\midrule
Pure Armijo, CIFAR-10/ResNet-18 & 8 & 56\% & $-0.87$ & $-0.60 \pm 0.07$ & $0.017$ & $-0.33 \pm 0.05$ & $0.330$ \\
Pure Armijo, Fashion-MNIST/CNN & 20 & 13\% & $-0.95$ & $-0.65 \pm 0.13$ & $< 10^{-4}$ & $-0.30 \pm 0.23$ & $0.066$ \\
Pure Armijo, Imagenette/ResNet-18 & 23 & 4\% & $-0.90$ & $-0.65 \pm 0.20$ & $< 10^{-5}$ & $-0.39 \pm 0.24$ & $0.017$ \\
\midrule
Adam $10^{-3}$, CIFAR-10/ResNet-18 & 40 & 0\% & $-0.86$ & $-0.50 \pm 0.05$ & $< 10^{-6}$ & $-0.37 \pm 0.07$ & $0.004$ \\
Adam $10^{-2}$ (near-EoS), CIFAR-10/ResNet-18 & 40 & 0\% & $-0.84$ & $-0.73 \pm 0.07$ & $< 10^{-18}$ & $-0.65 \pm 0.09$ & $< 10^{-12}$ \\
AdamW $10^{-3}$, CIFAR-10/ResNet-18 & 40 & 0\% & $-0.86$ & $-0.36 \pm 0.12$ & $< 10^{-3}$ & $-0.30 \pm 0.11$ & $0.023$ \\
SGD-m $5{\cdot}10^{-2}$, CIFAR-10/ResNet-18 & 40 & 0\% & $-0.76$ & $-0.53 \pm 0.13$ & $< 10^{-8}$ & $-0.48 \pm 0.22$ & $< 10^{-5}$ \\
Adam $10^{-3}$, Fashion-MNIST/CNN & 47 & 2\% & $-0.68$ & $-0.48 \pm 0.12$ & $< 10^{-6}$ & $-0.35 \pm 0.18$ & $0.004$ \\
Adam $10^{-2}$ (near-EoS), Fashion-MNIST/CNN & 45 & 5\% & $-0.54$ & $-0.54 \pm 0.06$ & $< 10^{-8}$ & $-0.43 \pm 0.19$ & $< 10^{-4}$ \\
AdamW $10^{-3}$, Fashion-MNIST/CNN & 47 & 2\% & $-0.67$ & $-0.40 \pm 0.11$ & $< 10^{-4}$ & $-0.29 \pm 0.16$ & $0.022$ \\
SGD-m $5{\cdot}10^{-2}$, Fashion-MNIST/CNN & 45 & 6\% & $-0.60$ & $-0.46 \pm 0.09$ & $< 10^{-6}$ & $-0.45 \pm 0.16$ & $< 10^{-4}$ \\
\bottomrule
\end{tabular}}
\end{table}

\begin{table}[htbp]
\centering
\caption{Trend-robustness check for the $\alpha$--$\lambda_1$
correlation of Table~\ref{tab:correlation} (per architecture, 3 seeds;
$n$ = probe points per run). Detrended: Pearson correlation of the
residuals of $\log\alpha$ and $\log\lambda_1$ after removing a
linear-in-step trend from each series; per-run $p$-values combined
across seeds with Fisher's method ($p_F$). First diff.: Pearson
correlation of probe-to-probe changes
$(\Delta\log\alpha, \Delta\log\lambda_1)$, immune to shared monotone
drift but dominated by the $\beta$-quantisation of $\alpha$ (see text).}
\label{tab:correlation-detrended}
\resizebox{\textwidth}{!}{%
\begin{tabular}{lcccccc}
\toprule
Variant & $n$ & raw $\bar{r}$ & detrended $\bar{r}$ & $p_F$ & first-diff $\bar{r}$ & $p_F$ \\
\midrule
Pure Armijo, CIFAR-10/ResNet-18 (3 seeds) & 19 & $-0.913 \pm 0.013$ & $-0.630 \pm 0.139$ & $< 10^{-5}$ & $-0.295 \pm 0.092$ & $0.165$ \\
Pure Armijo, Fashion-MNIST/CNN (3 seeds) & 23 & $-0.950 \pm 0.017$ & $-0.697 \pm 0.148$ & $< 10^{-9}$ & $-0.255 \pm 0.205$ & $0.083$ \\
Pure Armijo, Imagenette/ResNet-18 (3 seeds) & 24 & $-0.907 \pm 0.021$ & $-0.595 \pm 0.155$ & $< 10^{-6}$ & $-0.372 \pm 0.224$ & $0.004$ \\
\bottomrule
\end{tabular}}
\end{table}

What does the line search actually invert? The correlation with
$\lambda_1$ combines two facts: an exact, elementary one about
backtracking, and a separate alignment fact borrowed from the
Edge-of-Stability literature. Separating them shows what is ours and what
is not. Consider a local quadratic model with Hessian $H \succ 0$ and a
descent direction $d$, so that $g^\top d < 0$ and $d^\top H d > 0$. On
this model the second-order expansion is exact,
\[
  L(\theta + \alpha d) \;=\; L(\theta) + \alpha\, g^\top d
  + \tfrac{1}{2}\,\alpha^2\, d^\top H d ,
\]
so the Armijo condition
$L(\theta + \alpha d) \le L(\theta) + c\,\alpha\,g^\top d$ becomes
$\tfrac{1}{2}\,\alpha^2\, d^\top H d \le (1-c)\,\alpha\,(-g^\top d)$.
Dividing by $\alpha > 0$ and by $d^\top H d > 0$ shows that the line
search accepts a step exactly when
\[
  \alpha \;\le\; \alpha_{\text{acc}} \;:=\; \frac{2(1-c)}{q_d},
  \qquad
  q_d \;=\; \frac{d^\top H d}{-\,g^\top d},
\]
with the directional curvature $q_d$ introduced above. Backtracking from
$\alpha_{\max}=1$ by factor $\beta$ tests the candidates
$1, \beta, \beta^2, \dots$; if an Armijo-acceptable candidate is found
before the backtracking cap $K$ of Algorithm~\ref{alg:init-probe} is
exhausted, the search returns the first such candidate,
$\alpha = \beta^k$. Whenever the constraint binds, that is, the full step
$\alpha_{\max}=1$ is rejected and $k \ge 1$, the returned candidate
$\beta^k$ satisfies the acceptance condition while $\beta^{k-1}$ does
not, hence $\beta\,\alpha_{\text{acc}} < \alpha \le \alpha_{\text{acc}}$. Taking
reciprocals, the returned step obeys
\[
  \frac{q_d}{2(1-c)} \;\le\; \frac{1}{\alpha} \;<\; \frac{q_d}{2\beta(1-c)} .
\]
If the search instead exhausts the cap without an accepted candidate, it
returns the next, \emph{untested} candidate $\beta^{K}$, one rung below
the last rejected one, and only a one-sided statement survives: every
tested candidate was rejected, so $q_d > 2\beta(1-c)/\beta^{K}$ -- a
curvature floor with no upper bound, and on landscapes whose true reading
lies further below $\beta^{K}$ the saturated value still overestimates
the safe step. The deployed protocols handle this saturated case in two
ways, both listed in Table~\ref{tab:protocol}. The benchmark and
language-model deployments keep $\beta^{K}$ and cap conservatively at
$\kappa\beta^{K}$: there the saturated reading is empirically
batch-independent (Table~\ref{tab:probe-variance}), and where sharpening
is slow enough for the first re-probes to fire, the probationary window
of Section~\ref{sec:method} walks a too-high init reading down within a
few steps -- the language-model case. The cross-architecture suite
instead lets the \emph{init} probe extend the backtracking until a
candidate is accepted, and on landscapes whose true reading lies
\emph{below} the $\beta^{K}$ floor this extension is not cosmetic but
necessary: on the normalisation-free ResNet, which reads one to two
decades below $\beta^{K}$ (Appendix~\ref{app:kappa-perarch}), a cap
frozen at $\kappa\beta^{K}$ diverges even with the probationary window
(the re-probes saturate at the same floor, and divergence outruns
them), while the extended reading trains safely.

The quadratic model is needed only to name \emph{what} is bracketed, not
for the bracket itself. For a general twice continuously differentiable
loss, Taylor's theorem with integral remainder gives, at any tested step,
\[
  L(\theta + \alpha d) \;=\; L(\theta) + \alpha\, g^\top d
  + \alpha^2 \!\int_0^1\! (1-t)\; d^\top H(\theta + t\alpha d)\, d \;\mathrm{d}t ,
\]
so the Armijo condition at $\alpha$ is equivalent, with no approximation,
to
\[
  \alpha\,\bar q_d(\alpha) \;\le\; 2(1-c),
  \qquad
  \bar q_d(\alpha) \;:=\; \frac{2\int_0^1 (1-t)\, d^\top H(\theta + t\alpha d)\, d\,\mathrm{d}t}{-\,g^\top d} ,
\]
with the \emph{segment-averaged} directional curvature $\bar q_d$ over the
step actually taken; whenever $\bar q_d(\alpha) > 0$ this reads
$\alpha \le 2(1-c)/\bar q_d(\alpha)$, and a step with non-positive
segment-averaged curvature is always accepted. The accepted/rejected pair of candidates therefore brackets
exactly as above with $\bar q_d$ in place of $q_d$: acceptance of
$\beta^k$ and rejection of $\beta^{k-1}$ give
$1/\alpha \ge \bar q_d(\alpha)/(2(1-c))$ and
$1/\alpha < \bar q_d(\alpha/\beta)/(2\beta(1-c))$, which collapse to the
point statement whenever $\bar q_d$ is constant across the tested
segment -- exactly on the quadratic model, where $\bar q_d \equiv q_d$.
Two readings follow. First, what the probe inverts exactly on a
neural-network loss is the finite-step curvature along the segment a
step of that size would actually traverse; for a \emph{step-size}
safeguard this is the more relevant quantity, since what destabilises a
finite step is the curvature over that step, not at its starting point.
Second, the gap between $\bar q_d$ and the point quantity $q_d$ is an
empirical question, and it is precisely what the band statistics below
measure. That is, the line search inverts the segment-averaged
directional curvature $\bar q_d$ \emph{along the probed direction} --
exactly, for any smooth loss -- and the point curvature $q_d$ up to the
measured gap between the two, not $\lambda_1$; which curvature it
reports is then entirely a matter of which direction $d$ one probes, and
two choices matter.

\begin{figure}[htbp]
\centering
\includegraphics[width=0.62\linewidth]{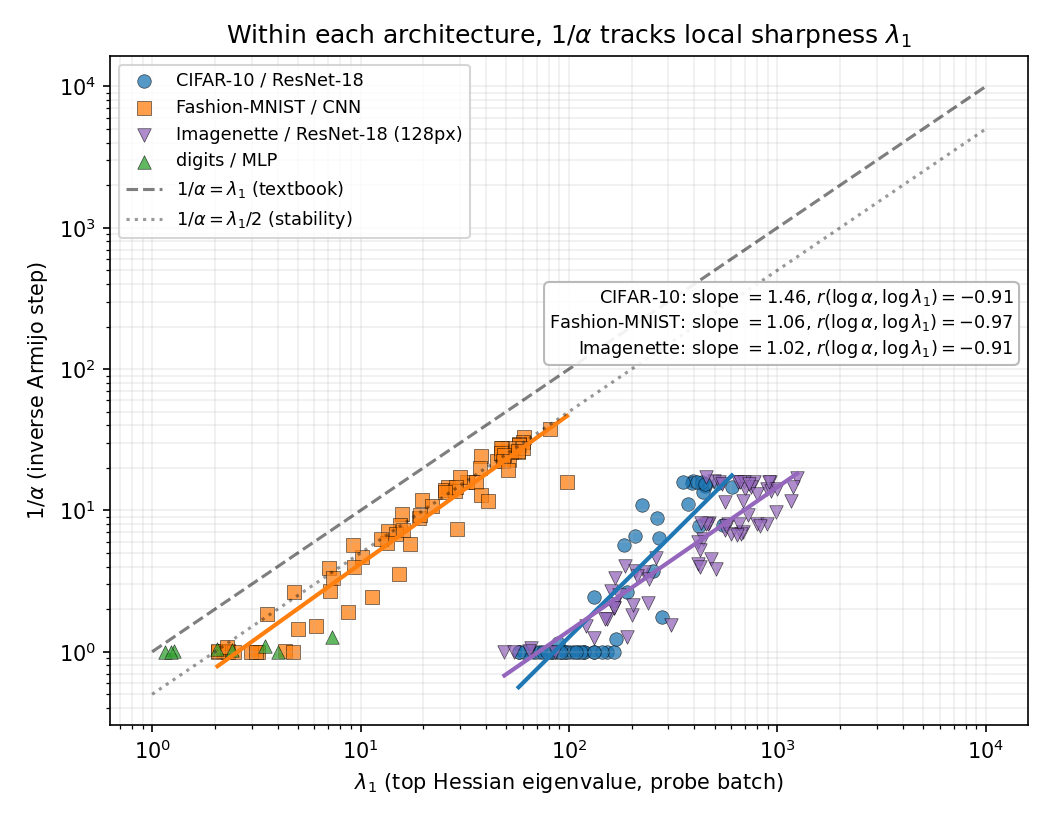}
\caption{Within each architecture, $1/\alpha$ rises with $\lambda_1$ on
log--log scales with fitted slopes $1.46$ (CIFAR-10), $1.06$
(Fashion-MNIST) and $1.02$ (Imagenette) and strong correlation
($r = -0.91$, $-0.97$ and $-0.91$ between
$\log\alpha$ and $\log\lambda_1$; the fits pool the three seeds of each
architecture, per-seed statistics in Table~\ref{tab:correlation}), but lies
a roughly constant factor \emph{below} the
textbook $1/\alpha = \lambda_1$ line (dashed), consistent with the step
inverting the directional curvature $q \le \lambda_1$ rather than
$\lambda_1$ itself. The correlation is a within-run effect, so
architectures are fitted separately and never pooled; digits (a small MLP
whose $\lambda_1$ spans under a decade) is shown only as context.}
\label{fig:scatter}
\end{figure}

The first is the \emph{raw-gradient probe}, $d=-g$, for which
$q_{-g} = g^\top H g/\|g\|^2$ is the Rayleigh quotient of $H$ along the
gradient; we write $q$ for $q_{-g}$ in the measurement experiments that
follow. This is the variant we use to expose the mechanism, because it
admits a clean link to the top eigenvalue. We test the band directly by
measuring $\alpha$ and $q_{-g}$ on the \emph{same} probe batch at each probe
(a single extra Hessian-vector product). Across the CIFAR-10 and
Fashion-MNIST mechanism runs, $73\%$ of the probe points at which the Armijo
constraint is active fall inside the predicted factor-$2$ band, with median
ratio $(1/\alpha)/(q_{-g}/(2(1-c))) = 1.27$, comfortably inside the predicted
$[1, 1/\beta] = [1, 2]$. The remaining $\approx 27\%$ sit outside the
band; by the exact finite-step statement above this is not slack in the
search itself but the measured gap between the segment-averaged curvature
$\bar q$ that Armijo inverts and the point quantity $q$ that the
Hessian-vector product measures (the mini-batch loss is only locally
quadratic, so the two differ over a finite tested segment), plus
probe-batch stochasticity. The distribution shows the gap is small: over
the same active probe points the ratio has p90 $= 1.92$, p95 $= 2.24$ and
maximum $3.57$ against a band edge of $2$, i.e.\ $97\%$ of active probes
lie within a factor of $2$ of the predicted band. The link to $\lambda_1$
is then the second,
EoS-side fact: we measure $q_{-g} \approx 0.63\,\lambda_1$ (median over both
architectures), i.e.\ the gradient carries a large but incomplete component
along the top eigenvector, so $1/\alpha \approx q_{-g}/2 \approx
0.3\,\lambda_1$, a strong linear relation with a constant offset, exactly
the parallel-but-below line seen in Figure~\ref{fig:scatter}. The novel,
exact part of the mechanism is $1/\alpha \propto q_{-g}$; the proportionality
$q_{-g} \propto \lambda_1$ is inherited from the Edge-of-Stability alignment
phenomenon~\citep{cohen2021eos,damian2023eos}.

The second choice is the one that makes the probe deployable. Adam does not
step along $-g$; it steps along the preconditioned direction
$d = -g/(\sqrt{v}+\epsilon)$, which at the first step reduces to
$-g/(|g|+\epsilon) \approx -\operatorname{sign}(g)$, the geometry the
probe uses at every probe time. Probing along
that direction measures $q_d$ in the geometry the optimiser actually moves
in: the curvature that matters is the curvature along the direction the
optimiser actually takes. The practical implication is: the raw-gradient probe
reports curvature in a coordinate system Adam never uses, so the safety
factor that converts a probe reading into a step cap is
architecture-specific, whereas the \emph{Adam-direction probe} yields a
single safety factor that transfers across networks unchanged. We establish
this next, and it is what lets one probe-derived rule guard Adam across all
the architectures we test.

Figure~\ref{fig:scatter} visualises this within each architecture:
$1/\alpha$ plotted against $\lambda_1$ on log--log scales follows a strong
linear trend with fitted slopes of order one ($1.46$ on CIFAR-10, $1.06$ on
Fashion-MNIST, $1.02$ on Imagenette), lying a roughly constant factor below the textbook
$1/\alpha = \lambda_1$ reference, consistent with the
$1/\alpha \approx 0.3\,\lambda_1$ estimate, a unit-slope law with a constant
offset that the fitted slopes bracket (close on Fashion-MNIST, somewhat
steeper on CIFAR-10).

\begin{figure}[htbp]
\centering
\includegraphics[width=\linewidth]{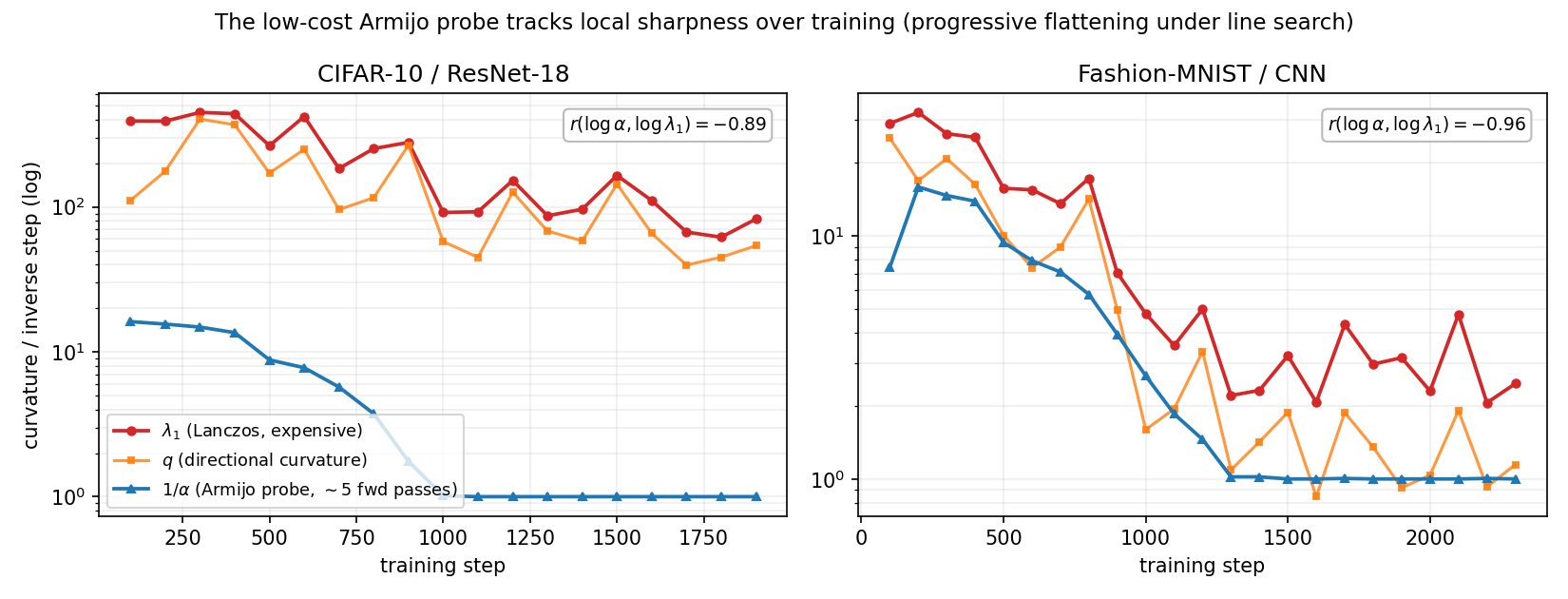}
\caption{The low-cost Armijo probe tracks local sharpness over training.
Per architecture (one run), the inverse line-search step $1/\alpha$
($\sim$5 forward passes) follows the directional curvature $q$ tightly and
the Lanczos top eigenvalue $\lambda_1$ up to a constant factor, across the
progressive flattening induced by the line search. $1/\alpha$ floors at $1$
once the full step is accepted (CIFAR-10, late training), bounding the
range it resolves.}
\label{fig:tracking}
\end{figure}

The correlation above is a static snapshot; in practice the probe is
useful because it tracks sharpness \emph{online}. Figure~\ref{fig:tracking}
overlays, against the training step, the expensive top eigenvalue
$\lambda_1$ (Lanczos), the directional curvature $q$, and the low-cost probe
reading $1/\alpha$, for one run per architecture. The probe follows
$\lambda_1$ throughout training (here a steady flattening, as the line
search drives the iterate into lower-curvature regions) for about five
forward passes per reading, against the $k$ Hessian--vector products (each
roughly two backward passes) that a Lanczos estimate of $\lambda_1$ needs.
The reading has a floor: once the landscape is flat enough that the full
step $\alpha = \alpha_{\max} = 1$ is accepted, $1/\alpha$ saturates at $1$
and no longer resolves curvature below that level (visible on CIFAR-10
after $\sim$1000 steps). Where the Armijo constraint is active, though, the
line-search step is a low-cost online sensor for local sharpness: a
practical way to watch Edge-of-Stability dynamics without Hessian machinery.

How faithfully does the step read curvature, and what controls the
fidelity? Table~\ref{tab:fidelity} characterises the probe as a sensor on
the Fashion-MNIST CNN, where $\lambda_1$ is cheap and reliable to estimate
(PyHessian/Lanczos) as a reference. Two
knobs matter. The backtracking factor $\beta$ sets the resolution: theory
places $1/\alpha$ in $[q/2(1-c),\, q/2\beta(1-c)]$, a band of width
$1/\beta$, and with the backtracking depth held fixed the reading tightens
exactly as predicted, from a median $(1/\alpha)/(q/2)$ of $1.8$ inside
$[1,2]$ at the default $\beta = 0.5$ to $1.07$ inside $[1,1.11]$ at
$\beta = 0.9$, at the cost of proportionally more evaluations. The probe
batch sets the noise: the relative standard deviation of the $\lambda_1$
and $q$ estimates over independent batches falls from $\approx 0.4$ at
batch $64$ to $\approx 0.1$ at batch $256$, so a few hundred examples
already give a stable reading. Across these settings the directional
curvature is a roughly constant fraction of the top eigenvalue,
$q/\lambda_1 \approx 0.6$, matching the alignment used above. The batch
dependence of the deployed one-shot reading itself is quantified in
Appendix~\ref{app:probe-variance}: re-run on $50$ different batches at
a frozen initialisation, the raw-gradient reading moves by at most one
backtracking rung on eleven of twelve model--seed combinations, and the
Adam-direction reading at initialisation is constant across batches
because it exhausts the backtracking cap.

\begin{table}[htbp]
\centering
\caption{Fidelity of the line-search reading (Fashion-MNIST/CNN, 8
checkpoints). With the backtracking depth matched across $\beta$, the
inverse step $1/\alpha$ reads the directional curvature $q$ within the
predicted band $[1, 1/\beta]$; a finer $\beta$ tightens the band towards $1$
at the cost of more evaluations. Probe-batch noise ($\lambda_1$/$q$
relative std $0.4\!\to\!0.1$ from batch $64\!\to\!256$) and the
$q \approx 0.6\,\lambda_1$ alignment are reported in the text.}
\label{tab:fidelity}
\begin{tabular}{lcc}
\toprule
backtracking factor $\beta$ & predicted band $[1,1/\beta]$ & median $(1/\alpha)/(q/2)$ (\% in band) \\
\midrule
$0.5$ (default) & $[1,\,2.00]$ & $1.79$ \ (86\%) \\
$0.7$           & $[1,\,1.43]$ & $1.23$ \ (86\%) \\
$0.9$           & $[1,\,1.11]$ & $1.07$ \ (100\%) \\
\bottomrule
\end{tabular}
\end{table}

This establishes the paper's measurement claim: the line search reads
local sharpness at low cost. We now turn to the practical pay-off: using
that reading, once at initialisation, to cap a mis-specified learning
rate.
Table~\ref{tab:rescue} reports test accuracy across the full lr range
$\{10^{-3}, 10^{-2}, 10^{-1}, 0.3, 1.0, 3.0\}$ for vanilla Adam and the
controllers, mean $\pm$ std over 5 seeds.

\begin{table}[htbp]
\centering
\caption{CIFAR-10 / ResNet-18, 10 epochs. Test accuracy at six initial
learning rates spanning more than three orders of magnitude. ``DIV.''\ =
all seeds diverged at the first update (loss $\to$ NaN); ``$x/n$ div.''\ =
partial divergence with the surviving-seed mean. Mean $\pm$ std over five
seeds; best per column within each block in bold.}
\label{tab:rescue}
\resizebox{\textwidth}{!}{%
\begin{tabular}{lcccccc}
\toprule
$\eta_{\text{init}}$ & $10^{-3}$ & $10^{-2}$ & $10^{-1}$ & $0.3$ & $1.0$ & $3.0$ \\
\midrule
Vanilla Adam     & $\mathbf{0.834 \pm 0.025}$ & $0.811 \pm 0.016$ & DIV.\           & DIV.\          & DIV.\          & DIV.\ \\
\textbf{Adam-InitOnly} & $0.833 \pm 0.023$ & $0.806 \pm 0.016$ & $\mathbf{0.807 \pm 0.023}$ & $\mathbf{0.799 \pm 0.026}$ & $\mathbf{0.794 \pm 0.038}$ & $\mathbf{0.796 \pm 0.023}$ \\
Adam-Watchdog    & $0.808 \pm 0.051$ & $\mathbf{0.816 \pm 0.020}$ & $0.797 \pm 0.010$ & $0.757 \pm 0.010$ & 0.233\,(2/5 div.) & 0.421\,(3/5 div.) \\
Adam-Tracker     & $0.803 \pm 0.023$ & $0.766 \pm 0.022$ & $0.756 \pm 0.027$ & $0.765 \pm 0.020$ & $0.761 \pm 0.026$ & $0.772 \pm 0.019$ \\
\midrule
\multicolumn{7}{l}{\textit{Direction-matched probe, fixed $\kappa = 2$ (no calibration)}} \\
\textbf{Adam-InitOnly (dir.)} & $\mathbf{0.839 \pm 0.023}$ & $\mathbf{0.823 \pm 0.015}$ & $\mathbf{0.828 \pm 0.016}$ & $\mathbf{0.824 \pm 0.022}$ & $\mathbf{0.817 \pm 0.018}$ & $\mathbf{0.823 \pm 0.020}$ \\
\midrule
\multicolumn{7}{l}{\textit{Transfer check: AdamW base optimiser}} \\
Vanilla AdamW & $\mathbf{0.843 \pm 0.013}$ & $0.809 \pm 0.020$ & DIV.\ & DIV.\ & DIV.\ & DIV.\ \\
\textbf{AdamW-InitOnly (dir.)} & $0.828 \pm 0.026$ & $\mathbf{0.823 \pm 0.014}$ & $\mathbf{0.827 \pm 0.017}$ & $\mathbf{0.816 \pm 0.016}$ & $\mathbf{0.818 \pm 0.017}$ & $\mathbf{0.826 \pm 0.018}$ \\
\midrule
\multicolumn{7}{l}{\textit{Parameter-free / schedule-free baselines (no $\kappa$; $\eta$ swept like the rest)}} \\
Schedule-Free AdamW & $\mathbf{0.865 \pm 0.005}$ & $\mathbf{0.864 \pm 0.006}$ & DIV.\           & DIV.\          & DIV.\          & DIV.\ \\
Prodigy          & $0.100 \pm 0.005$ & $0.529 \pm 0.003$ & $\mathbf{0.768 \pm 0.006}$ & $\mathbf{0.795 \pm 0.011}$ & $\mathbf{0.813 \pm 0.010}$ & $\mathbf{0.827 \pm 0.018}$ \\
\bottomrule
\end{tabular}}
\end{table}

\begin{figure}[htbp]
\centering
\includegraphics[width=\linewidth]{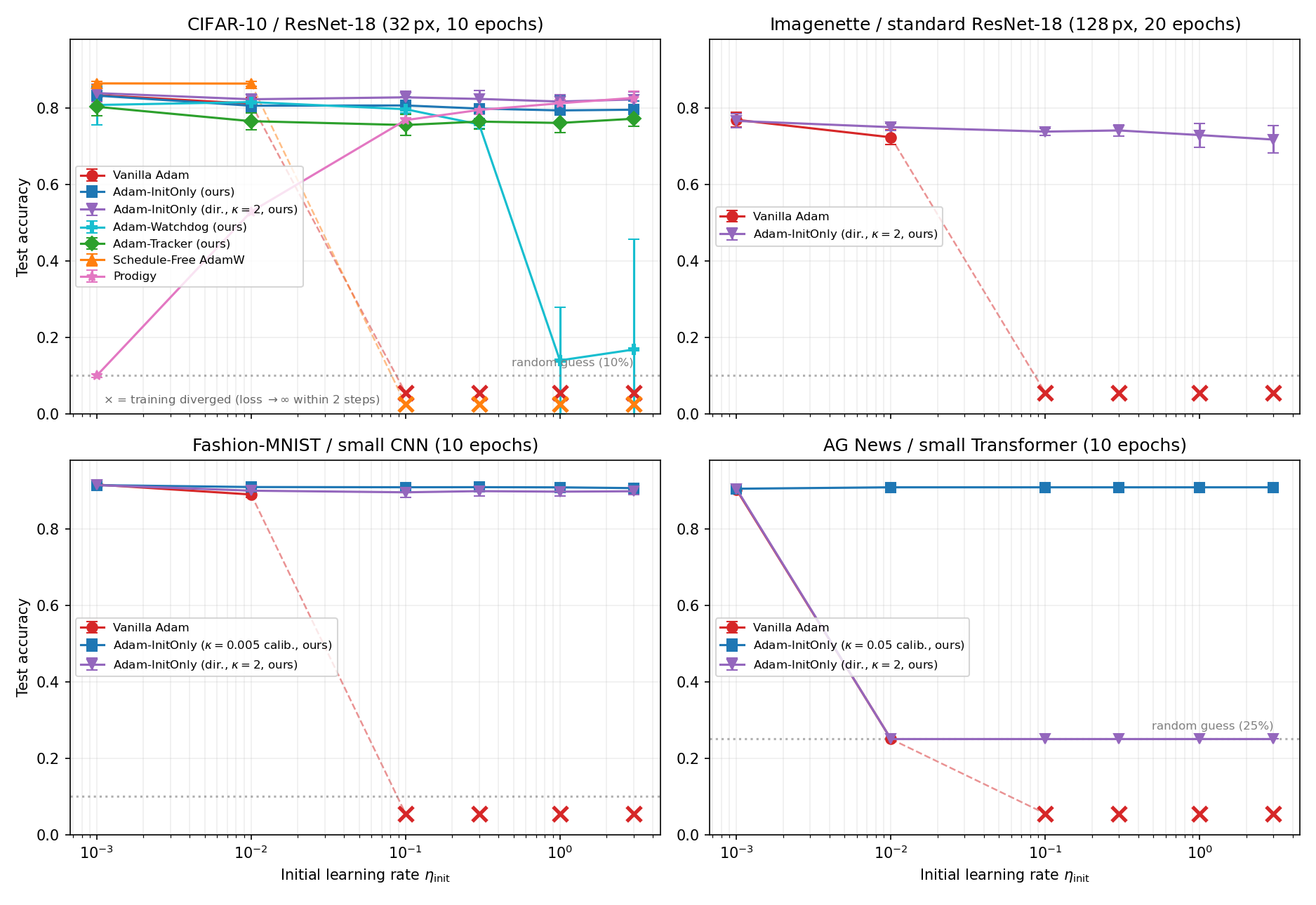}
\caption{Test accuracy vs.\ initial learning rate, mean$\pm$std over five
seeds; ``$\times$'' markers at the bottom mark runs that diverged (loss
$\to\infty$ within two steps), with dashed lines connecting each
method's last non-diverging point to its divergence markers; dotted
grey lines mark random-guess accuracy. \emph{Top left:}
CIFAR-10/ResNet-18, all methods. Vanilla Adam (red) and Schedule-Free
AdamW (orange) diverge for $\eta \ge 10^{-1}$, the parameter-free
baselines each cover only
part of the range, and the calibrated Adam-InitOnly (blue) and the
calibration-free direction-matched variant at the fixed $\kappa = 2$
(purple) stay near-optimal across all six rates. \emph{Top right:} the
real-resolution Imagenette control (standard full-stem ResNet-18 at
$128$px) repeats the pattern with the same fixed $\kappa = 2$.
\emph{Bottom left:} Fashion-MNIST, where the calibrated raw-gradient
probe ($\kappa = 0.005$) and the calibration-free $\kappa = 2$ coincide.
\emph{Bottom right:} AG News makes the limit of the calibration-free
recipe visible: the fixed $\kappa = 2$ still prevents every divergence
but settles on the random-guess plateau for $\eta \ge 10^{-2}$, while
the one-minute calibration ($\kappa = 0.05$) recovers full accuracy.}
\label{fig:rescue-headline}
\end{figure}

Vanilla Adam is competitive at the optimal $\eta = 10^{-3}$ ($0.834$)
but diverges in the very first step at any $\eta \ge 10^{-1}$; the
watchdog absorbs one further decade of misspecification before failing,
and the tracker survives the whole range at a persistent accuracy cost
(Figure~\ref{fig:rescue-headline}). The clearest comparison is the simplest controller. Adam-InitOnly, which performs only the calibration probe
before the first Adam step and then leaves Adam entirely alone, matches
vanilla Adam at the optimal $\eta$ ($0.833$ vs $0.834$, within seed
noise) and stays in $[0.79, 0.83]$ across the full range. It matches or
exceeds Adam-Tracker at every learning rate, has no periodic compute
overhead, and, as the transfer experiments below show, is the only
controller whose settings carry over to other architectures. We
recommend Adam-InitOnly as the practical default.

The fair comparison is not vanilla Adam at an arbitrary learning
rate (which no practitioner would deliberately choose) but vanilla
Adam whose $\eta$ has been \emph{tuned}, typically by a learning-rate
sweep or range test~\citep{smith2017cyclical}. Adam-InitOnly delivers
essentially that tuned accuracy without the search: its worst cell over
the entire range ($0.794$ at $\eta = 1.0$) is within five percent of
optimally-tuned vanilla Adam's best ($0.834$ at $\eta = 10^{-3}$), and it
reaches this from a single probe at \emph{every} learning rate. We make this
comparison concrete in Table~\ref{tab:rangetest}, on all four benchmarks:
against an actual learning-rate range test~\citep{smith2017cyclical} that
ramps $\eta$ exponentially over $\sim\!2$ epochs and then trains at a tenth
of the loss-minimising rate, the direction-matched probe (handed a
deliberately hundredfold-too-large $\eta = 0.1$ to cap) matches the range
test's safety exactly: zero divergences in every run on both sides.
Accuracy separates the two roles. On CIFAR-10 the probe is slightly ahead
($0.828$ vs $0.810$); on Fashion-MNIST and Imagenette the range test's
actual search buys two to three points ($0.917$ vs $0.896$ and $0.770$ vs
$0.738$); and on AG News, where the fixed cap lands above the productive
range, the difference is qualitative: the range test finds
$\eta^{\ast} \approx 10^{-3}$ and trains to $0.904$ while the capped run
survives at chance, the limit discussed in
Section~\ref{sec:results-fashion}. The computational cost difference is substantial. Counted
architecture-invariantly in optimiser-step-equivalent forward-and-backward
passes, the ramp costs ten to twenty percent of the training budget,
whereas the probe adds one such pass plus eight forward-only evaluations,
below one percent of every budget here. The probe is therefore best read
not as beating the range test but as buying its \emph{safety} role, not
its search, at a fraction of the cost.

\begin{table}[htbp]
\centering
\caption{A real learning-rate range test versus the one-shot probe on all
four benchmarks, mean$\pm$std over 5 seeds. The range test ramps $\eta$
exponentially over $\sim\!2$ epochs and trains at a tenth of the
loss-minimising rate; the direction-matched probe ($\kappa = 2$) is handed
a deliberately hundredfold-too-large $\eta = 0.1$ and caps it once at
initialisation. Extra cost is counted in forward-and-backward passes (fb)
and stated relative to the training budget; the probe adds one such pass
plus eight forward-only evaluations, below one percent of every budget.
No run of either method diverged. On AG News the capped run survives only
at chance accuracy (see text).}
\label{tab:rangetest}
\resizebox{\textwidth}{!}{%
\begin{tabular}{lccccc}
\toprule
 & \multicolumn{3}{c}{LR range test (Smith) $\to$ train} & \multicolumn{2}{c}{Adam-InitOnly (dir.) @ $\eta = 0.1$} \\
\cmidrule(lr){2-4}\cmidrule(lr){5-6}
Benchmark & $\eta^{\ast}$ & test acc & extra cost & $\eta_{\text{cap}}$ & test acc \\
\midrule
CIFAR-10/ResNet-18 & $0.0216$ & $0.810 \pm 0.015$ & $392$\,fb ($20\%$) & $0.0078$ & $0.828 \pm 0.016$ \\
Fashion-MNIST/CNN & $0.0030$ & $0.917 \pm 0.003$ & $366$\,fb ($16\%$) & $0.0094$ & $0.896 \pm 0.014$ \\
AG News/Transformer & $0.0009$ & $0.904 \pm 0.003$ & $847$\,fb ($18\%$) & $0.0078$ & $0.250 \pm 0.000$ \\
Imagenette/ResNet-18 & $0.0017$ & $0.770 \pm 0.009$ & $74$\,fb ($10\%$) & $0.0078$ & $0.738 \pm 0.009$ \\
\bottomrule
\end{tabular}}
\end{table}

An even simpler practitioner heuristic guards against the same failure
mode: retry on divergence. Start at the chosen rate, watch for
divergence, and restart from scratch one grid rung lower until training
survives. Table~\ref{tab:retry} quantifies this baseline on all four
benchmarks by replaying the recorded grid runs from the worst-case
start $\eta_0 = 3.0$: each diverged attempt is charged the optimiser
steps until the divergence criterion of Section~\ref{sec:setup} fires,
and the run then restarts one rung lower. Because divergence at these
rates is detected within a few dozen steps, retrying is genuinely
competitive on this suite: it always terminates at the same largest
stable grid rate, its total detection cost stays below four percent of
one training run, and its final accuracy sits within about one
percentage point of the probe's. The probe does not claim to reach
rates a restart ladder cannot reach; it claims to remove the failure
mode proactively, and the differences are operational. The probe pays
its one percent before the first step, needs no divergence detector, no
restart logic and no candidate grid, keeps the run deterministic and
single-shot, and returns a curvature reading as a by-product. The retry
ladder is reactive, its cost scales with the detection latency (a few
steps here, but a diverged warmup is far more expensive at scales where
checkpoint cadence and cluster scheduling dominate), and its detector
threshold is itself a small calibration. On AG News the table also
shows a limit the two strategies share: the largest non-diverging rate
still trains at chance accuracy, so surviving is not the same as
learning, and neither a cap nor a restart ladder replaces accuracy
monitoring (Section~\ref{sec:results-fashion}).

\begin{table}[htbp]
\centering
\caption{Retry-on-divergence versus the one-shot probe, from the
worst-case start $\eta_0 = 3.0$, mean over 5 seeds, simulated by
replaying the recorded grid runs: whenever a run diverges, it restarts
one rung lower on the ladder
$3.0 \to 1.0 \to 0.3 \to 0.1 \to 0.01 \to 0.001$. Wasted steps count
optimiser steps until the divergence criterion fires, summed over
restarts ($\le$ marks upper bounds from the 20-step logging grid where
the exact detection step was not logged); overhead is relative to one
full training run. Adam-InitOnly (dir.) is the direction-matched probe
at $\kappa = 2$ handed the same $\eta_0 = 3.0$. On AG News both methods
survive at the largest stable grid rate yet train at chance accuracy
(see text).}
\label{tab:retry}
\resizebox{\textwidth}{!}{%
\begin{tabular}{lccccc}
\toprule
Benchmark & restarts & wasted steps & overhead & retry acc & Adam-InitOnly (dir.) acc \\
\midrule
CIFAR-10/ResNet-18 & 4 & $\le 37$ & $\le 1.9\%$ & $0.811 \pm 0.016$ & $0.823 \pm 0.020$ \\
Fashion-MNIST/CNN & 4 & $\le 80$ & $\le 3.4\%$ & $0.890 \pm 0.008$ & $0.898 \pm 0.009$ \\
AG News/Transformer & 4 & $11$ & $0.2\%$ & $0.250 \pm 0.000$ & $0.250 \pm 0.000$ \\
Imagenette/ResNet-18 & 4 & $8$ & $1.1\%$ & $0.724 \pm 0.020$ & $0.718 \pm 0.036$ \\
\bottomrule
\end{tabular}}
\end{table}

Modern parameter-free optimisers do not remove the need for this
safeguard. The last two rows of Table~\ref{tab:rescue} report two recent
parameter-free / schedule-free optimisers run across the same grid:
Schedule-Free AdamW~\citep{defazio2024scheduleFree} and
Prodigy~\citep{mishchenko2023prodigy}.\footnote{Prodigy is normally run
at its default multiplier $\eta = 1.0$, where it indeed performs well
here; we sweep the multiplier like every other method because our
question is precisely how each method behaves when the user-specified
scale is wrong.} Neither covers the full range.
Schedule-Free AdamW is strongest of all at small $\eta$ ($0.865$/$0.864$
at $\eta \le 10^{-2}$, above every other method) but still diverges in
the first step for any $\eta \ge 10^{-1}$, exactly the failure our probe
targets. Prodigy never diverges, but its internal step-size estimate
starts far too small at $\eta = 10^{-3}$ (test accuracy at random-guess
level) and only becomes competitive once $\eta \ge 10^{-1}$, where it is
in fact the best method ($0.813$/$0.827$ at $\eta = 1.0$/$3.0$). Each
parameter-free optimiser thus excels in a different part of the range but
collapses at the opposite end, whereas Adam-InitOnly stays within about
four percentage points of its own optimum across the entire
$10^{-3}$--$3.0$ range with no divergence; the probe is a calibrate-once safety check
for Adam, not a competing optimiser, and these results are complementary
rather than contradictory.

The direction-matched block of Table~\ref{tab:rescue} removes even the
one-time calibration: at the single fixed $\kappa = 2$, the
direction-matched probe matches the calibrated raw-gradient variant
across the entire grid ($0.82$--$0.84$, zero divergences), with no
per-architecture sweep at all. The AdamW block repeats the experiment
with the base optimiser swapped: vanilla AdamW diverges from
$\eta = 0.1$ exactly like Adam, and the identical probe at the identical
$\kappa$ restores the full range.

\begin{figure}[htbp]
\centering
\includegraphics[width=0.78\linewidth]{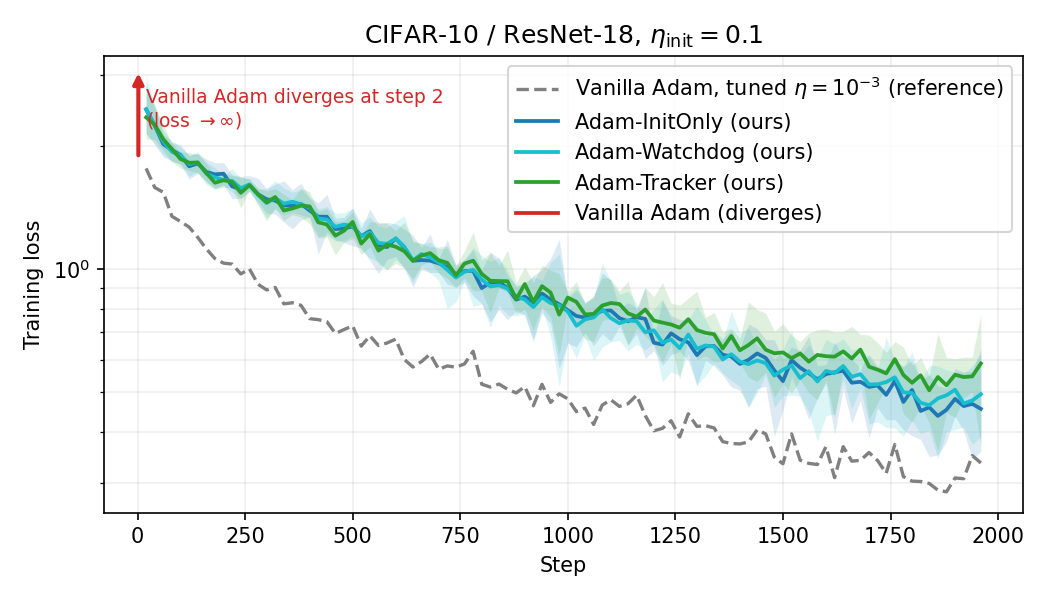}
\caption{Training loss at the misspecified $\eta_{\text{init}} = 0.1$
(mean over 5 seeds, shaded min--max). The grey dashed reference is vanilla
Adam at its well-tuned $\eta = 10^{-3}$, which trains normally: Adam is
not failing at CIFAR-10, only at this $100\times$-too-large rate. At
$\eta = 0.1$ vanilla Adam instead diverges at step~2 (loss $\to\infty$, off
chart), whereas Adam-InitOnly and the two periodic controllers train fine
and approach the tuned reference, all from the same bad $\eta_{\text{init}}$.}
\label{fig:rescue-loss}
\end{figure}

\begin{figure}[htbp]
\centering
\includegraphics[width=0.78\linewidth]{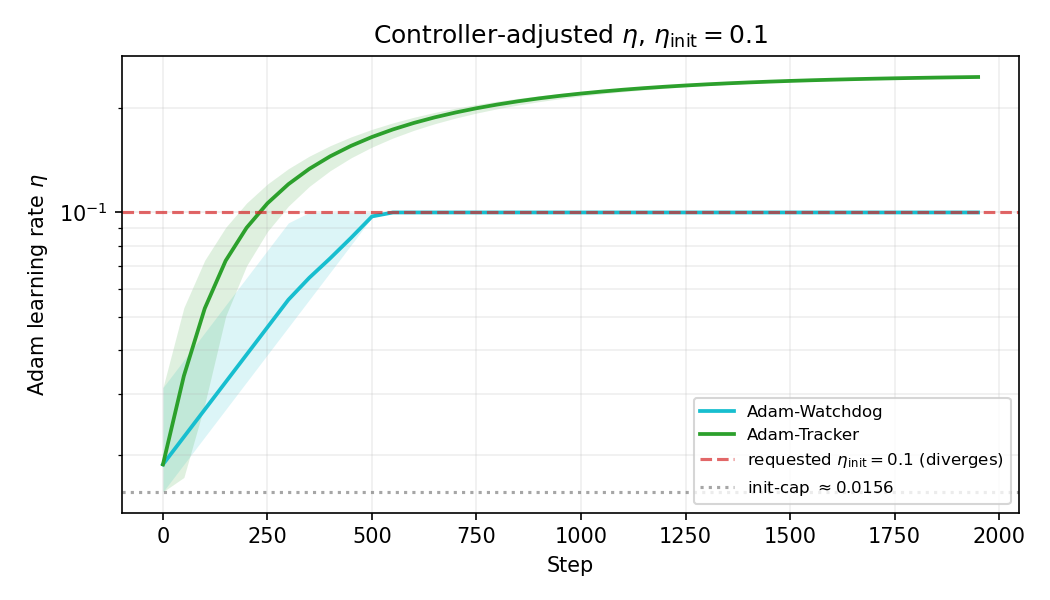}
\caption{Controller-adjusted $\eta$ at $\eta_{\text{init}} = 0.1$ (mean
over 5 seeds, shaded min--max). The init probe caps the requested
$\eta = 0.1$ to $\approx 0.016$ before training (dotted); the watchdog
then relaxes $\eta$ back to the user's $0.1$ ceiling, while the tracker
follows the probed step size up to $\approx 0.25$.}
\label{fig:lr-trace}
\end{figure}

Figure~\ref{fig:rescue-loss} shows the $\eta = 10^{-1}$ case
explicitly: vanilla Adam's loss explodes within two steps and is
irrecoverable, while Adam-InitOnly and the two periodic controllers all
train normally. Figure~\ref{fig:lr-trace} shows what the controllers
actually do with $\eta$: the non-divergent behaviour is established by
the initial cap, which caps the requested $\eta = 0.1$ down to
$\approx 0.016$ before the first update. From there the two periodic
controllers \emph{relax} the rate rather than throttle it further: the
watchdog climbs back to the user's $0.1$ ceiling and holds, while the
tracker follows the probed step size up to $\approx 0.25$. The divergence prevention is
therefore attributable to the one-shot cap, not to continued shrinking
during training, which is precisely why the init probe alone suffices.

\phantomsection\label{sec:results-baselines}
A reasonable concern is that the divergence-prevention effect is also achieved by
simpler standard tricks. Table~\ref{tab:baselines} compares the
controllers against gradient norm clipping ($\|g\| \le 1.0$) and a
200-step linear warmup at the two diagnostic learning rates
$\eta = 10^{-1}$ and $\eta = 1.0$.

\begin{table}[htbp]
\centering
\caption{Comparison of the probe controllers with low-cost standard
baselines (gradient-norm clipping, linear warmup) at a well-tuned
($\eta = 10^{-3}$) and two misspecified ($\eta = 10^{-1}, 1.0$) initial
learning rates; CIFAR-10/ResNet-18, mean$\pm$std over five seeds; best per
column within each block in bold.
``DIV.''\ = all seeds diverged at the first update;
``$x/n$ div.''\ = partial divergence (surviving-seed mean).}
\label{tab:baselines}
\begin{tabular}{lccc}
\toprule
                 & $\eta_{\text{init}} = 10^{-3}$ & $\eta_{\text{init}} = 10^{-1}$ & $\eta_{\text{init}} = 1.0$ \\
\midrule
\multicolumn{4}{l}{\textit{Vanilla Adam and standard defences}} \\
Vanilla Adam     & $0.834 \pm 0.025$ & DIV.\                  & DIV.\ \\
Adam + grad-clip ($\|g\| \le 1$) & $\mathbf{0.836 \pm 0.021}$ & DIV.\ & DIV.\ \\
Adam + 200-step warmup     & $0.835 \pm 0.015$ & $\mathbf{0.809 \pm 0.028}$ & $0.10$\textsuperscript{*}\,(3/5 div.) \\
\midrule
\multicolumn{4}{l}{\textit{Probe-based controllers (ours)}} \\
\textbf{Adam-InitOnly}     & $\mathbf{0.833 \pm 0.023}$ & $\mathbf{0.807 \pm 0.023}$ & $\mathbf{0.794 \pm 0.038}$ \\
Adam-Watchdog    & $0.808 \pm 0.051$      & $0.797 \pm 0.010$      & $0.233$\,(2/5 div.) \\
Adam-Tracker     & $0.803 \pm 0.023$      & $0.756 \pm 0.027$      & $0.761 \pm 0.026$ \\
\bottomrule
\end{tabular}
\\[2pt]
{\footnotesize\textsuperscript{*}\,Surviving seeds failed to learn;
test accuracy at random-guess level.}
\end{table}

The two low-cost baselines are revealing in different ways. Global
gradient-norm clipping at $\|g\|\le 1$ leaves vanilla Adam essentially
unchanged at well-tuned learning rates ($0.836 \pm 0.021$ at
$\eta = 10^{-3}$ against vanilla Adam's $0.834 \pm 0.025$, identical
within seed noise), so the clip neither helps nor hurts there. At
misspecified learning rates it provides
no divergence prevention at all: Adam still diverges in the first step at $\eta =
10^{-1}$ and at $\eta = 1.0$. The reason is structural rather than
numerical. Adam's update is $\eta\, m_t/(\sqrt{v_t} + \varepsilon)$, and
during the first few steps $v_t$ is initialised at zero and the
$1/\sqrt{v_t}$ factor amplifies even a clipped gradient by orders of
magnitude; clipping the gradient \emph{before} this amplification has
limited effect on the resulting parameter update. Linear warmup performs
better but only partially: it absorbs $\eta = 10^{-1}$ at accuracy on par
with the init probe ($0.809 \pm 0.028$ vs $0.807 \pm 0.023$), but at
$\eta = 1.0$ the warmup target is itself unstable, with three of five
seeds diverging and the surviving seeds settling at the random-guess
level. Of the probe-based
controllers, both Adam-InitOnly and Adam-Tracker survive every tested learning
rate up to $\eta = 3.0$ (Table~\ref{tab:rescue}), whereas Adam-Watchdog fails
from $\eta = 1.0$ on; Adam-InitOnly does so at the higher accuracy and, as we
show next, is the only one whose single setting also transfers to a second
architecture.

Table~\ref{tab:lowcost-cross} repeats this comparison across all four
benchmarks, at the framework default $\eta = 10^{-3}$ and at the
hundredfold misspecification $\eta = 10^{-1}$, with each method's extra
cost counted architecture-invariantly in additional passes. The CIFAR-10
pattern generalises. Gradient clipping never prevents divergence at a misspecified rate
on any benchmark. Warmup, the strongest conventional defence, is
benchmark-dependent: it absorbs $\eta = 10^{-1}$ on CIFAR-10 and
Imagenette at reduced accuracy, but all five seeds diverge on
Fashion-MNIST and the AG News survivors sit at chance. The parameter-free
optimisers repeat their complementary ends everywhere: Schedule-Free
AdamW is strong at the default rate (the best method in the table on
CIFAR-10 and Imagenette) yet diverges at $10^{-1}$ on every benchmark,
while Prodigy survives both rates on all four benchmarks but starts at or
below chance at the default multiplier on all four. The direction-matched
probe at the fixed $\kappa = 2$ is the only method that combines zero
divergences with near-default accuracy at both rates on the three vision
benchmarks; on AG News it inherits the chance-level limit that every
capped method shares (Section~\ref{sec:results-fashion}).

\begin{table}[htbp]
\centering
\caption{Low-cost safeguards across all four benchmarks: test accuracy at
the framework default $\eta = 10^{-3}$ and at the hundredfold-too-large
$\eta = 10^{-1}$, mean$\pm$std over 5 seeds, identical budgets and
divergence rule as the grid tables. ``DIV.''\ = all seeds diverged;
$x/n$~div.\ = partial divergence (surviving-seed mean). Extra cost is
counted in additional forward-and-backward (fb) and forward-only (f)
passes through the network, which is architecture- and
hardware-invariant; the parameter-free optimisers add optimiser-internal
arithmetic rather than extra passes. Every method shown is cheap, and
the differences lie in coverage. Chance accuracy is $0.1$ on
the vision benchmarks and $0.25$ on AG News.}
\label{tab:lowcost-cross}
\resizebox{\textwidth}{!}{%
\begin{tabular}{lccccccccc}
\toprule
 & & \multicolumn{2}{c}{CIFAR-10/RN18} & \multicolumn{2}{c}{Fashion/CNN} & \multicolumn{2}{c}{AG News/Transf.} & \multicolumn{2}{c}{Imagenette/RN18} \\
\cmidrule(lr){3-4}\cmidrule(lr){5-6}\cmidrule(lr){7-8}\cmidrule(lr){9-10}
Method & extra passes & $10^{-3}$ & $10^{-1}$ & $10^{-3}$ & $10^{-1}$ & $10^{-3}$ & $10^{-1}$ & $10^{-3}$ & $10^{-1}$ \\
\midrule
Vanilla Adam & 0 & $0.834 \pm 0.025$ & DIV. & $0.916 \pm 0.003$ & DIV. & $0.902 \pm 0.003$ & DIV. & $0.769 \pm 0.019$ & DIV. \\
Adam + grad-clip ($\|g\| \le 1$) & 0 & $0.836 \pm 0.021$ & DIV. & $0.917 \pm 0.002$ & DIV. & $0.905 \pm 0.002$ & DIV. & $0.760 \pm 0.029$ & DIV. \\
Adam + 200-step warmup & 0 & $0.835 \pm 0.015$ & $0.809 \pm 0.028$ & $0.913 \pm 0.004$ & DIV. & $0.902 \pm 0.002$ & $0.250 \pm 0.000$ & $0.759 \pm 0.027$ & $0.661 \pm 0.044$ \\
Schedule-Free AdamW & 0 & $0.865 \pm 0.005$ & DIV. & $0.914 \pm 0.001$ & DIV. & $0.903 \pm 0.003$ & $0.250$\,(3/5 div.) & $0.813 \pm 0.004$ & DIV. \\
Prodigy & 0 & $0.100 \pm 0.005$ & $0.768 \pm 0.006$ & $0.102 \pm 0.046$ & $0.915 \pm 0.004$ & $0.232 \pm 0.007$ & $0.906 \pm 0.002$ & $0.103 \pm 0.010$ & $0.725 \pm 0.013$ \\
Adam-InitOnly (dir., $\kappa=2$) & $1$\,fb$+8$\,f & $0.839 \pm 0.023$ & $0.828 \pm 0.016$ & $0.914 \pm 0.003$ & $0.896 \pm 0.014$ & $0.904 \pm 0.003$ & $0.250 \pm 0.000$ & $0.767 \pm 0.018$ & $0.738 \pm 0.009$ \\
\bottomrule
\end{tabular}}
\end{table}

\phantomsection\label{sec:results-fashion}
We next evaluate whether the controller settings transfer beyond
CIFAR-10/ResNet-18. To test this, we ran the same
sweep on the Fashion-MNIST CNN, with the same
default $\kappa = 0.25$ and watchdog thresholds calibrated on
CIFAR-10. Five seeds per cell. The result is reported in the upper block of
Table~\ref{tab:fashion-rescue}, alongside the calibrated version
discussed below.

Inspecting the controller logs reveals two distinct failures:
\emph{(i)} On Fashion-MNIST the probe saturates: the Armijo condition
already holds at the full step, so $\bar\alpha_{\text{init}} = 1$,
unlike on CIFAR-10, where backtracking engages and returns
$\bar\alpha_{\text{init}} \approx 0.06$. At the same time the largest
\emph{stable} Adam learning rate on Fashion-MNIST/small-CNN is only
$\approx 0.01$ (cf.\ vanilla Adam's divergence at $\eta = 0.1$), so the
stable fraction of the probed step is $\approx 0.01$ here against
$\approx 0.3$ on CIFAR-10; the conversion factor $\kappa$ must therefore
absorb a gap of well over an order of magnitude. The default
safety factor $\kappa = 0.25$ caps $\eta$ at $0.25$, far above the
stable range: at $\eta = 0.1$ the cap does not engage at all, and at the
larger requested rates it does engage but only reduces them to $0.25$,
which is itself still unstable, so the init-probe divergence-prevention mechanism cannot help here. \emph{(ii)} On the small CNN, sharpness rises much
faster in the first hundred steps (progressive sharpening is more
violent on a small network), so $\alpha$ collapses from $1$ to
$\approx 0.008$ within 100 steps. Watchdog interprets this as a
divergence signal and shrinks $\eta$ from $0.001$ to its floor
$5\times 10^{-5}$, well below the optimum.

The fix follows directly from this diagnosis, and it makes the recipe
simpler rather than more complex.
We re-ran the same Fashion-MNIST sweep with $\kappa$ recalibrated
to the architecture: the largest stable vanilla-Adam learning rate is
$\approx 10^{-2}$ (cf.\ the upper block of Table~\ref{tab:fashion-rescue}), and
$\bar\alpha_{\text{init}} \approx 1$, so we pick
$\kappa = 0.005$ (half of the empirical stability boundary).

\begin{table}[htbp]
\centering
\caption{Fashion-MNIST divergence prevention: uncalibrated $\kappa = 0.25$ versus
calibrated $\kappa = 0.005$.
Mean$\pm$std over 5 seeds; best per column within each block in bold.
``DIV.''\ = all 5 seeds diverged;
``$(x/n)$''\ = $x$ of $n$ seeds diverged, surviving-seed mean shown.
$^*$ surviving seeds learned nothing.}
\label{tab:fashion-rescue}
\resizebox{\textwidth}{!}{%
\footnotesize
\setlength{\tabcolsep}{2pt}%
\begin{tabular}{lcccccc}
\toprule
$\eta_{\text{init}}$ & $10^{-3}$ & $10^{-2}$ & $10^{-1}$ & $0.3$ & $1.0$ & $3.0$ \\
\midrule
\multicolumn{7}{l}{\textit{Uncalibrated, $\kappa = 0.25$ (CIFAR-10 default)}} \\
Vanilla Adam     & $0.916 \pm 0.003$ & $0.890 \pm 0.008$ & DIV.\ & DIV.\ & DIV.\ & DIV.\ \\
Adam + grad-clip & $\mathbf{0.917 \pm 0.002}$ & $0.899 \pm 0.004$ & DIV.\ & DIV.\ & DIV.\ & DIV.\ \\
Adam + warmup    & $0.913 \pm 0.004$ & $\mathbf{0.903 \pm 0.003}$ & DIV.\ & 0.10$^*$\,(2/5) & 0.10$^*$ & DIV.\ \\
Adam-Watchdog    & $0.878 \pm 0.002$ & $0.889 \pm 0.013$ & DIV.\ & DIV.\ & DIV.\ & DIV.\ \\
Adam-Tracker     & 0.81\,(3/5) & 0.10$^*$\,(4/5) & DIV.\ & DIV.\ & DIV.\ & DIV.\ \\
\midrule
\multicolumn{7}{l}{\textit{Calibrated, $\kappa = 0.005$ (one extra divergence sweep, $<\!1$ minute)}} \\
\textbf{Adam-InitOnly} & $\mathbf{0.914 \pm 0.005}$ & $\mathbf{0.909 \pm 0.004}$ & $\mathbf{0.909 \pm 0.004}$ & $\mathbf{0.909 \pm 0.004}$ & $\mathbf{0.908 \pm 0.003}$ & $\mathbf{0.907 \pm 0.003}$ \\
Adam-Watchdog    & $0.878 \pm 0.002$ & $\mathbf{0.909 \pm 0.005}$ & $\mathbf{0.909 \pm 0.004}$ & $0.895 \pm 0.004$ & 0.65\,(1/5) & 0.10$^*$\,(4/5) \\
Adam-Tracker     & DIV.\ & 0.46\,(3/5) & 0.10$^*$\,(4/5) & 0.84\,(4/5) & 0.10$^*$\,(3/5) & DIV.\ \\
\midrule
\multicolumn{7}{l}{\textit{Direction-matched probe, fixed $\kappa = 2$ (no calibration)}} \\
\textbf{Adam-InitOnly (dir.)} & $\mathbf{0.914 \pm 0.004}$ & $\mathbf{0.900 \pm 0.005}$ & $\mathbf{0.896 \pm 0.015}$ & $\mathbf{0.899 \pm 0.014}$ & $\mathbf{0.897 \pm 0.012}$ & $\mathbf{0.898 \pm 0.010}$ \\
\midrule
\multicolumn{7}{l}{\textit{Parameter-free / schedule-free baselines (no $\kappa$)}} \\
Schedule-Free AdamW & $\mathbf{0.914 \pm 0.001}$ & $\mathbf{0.920 \pm 0.002}$ & DIV.\ & DIV.\ & DIV.\ & DIV.\ \\
Prodigy          & $0.102 \pm 0.046$ & $0.876 \pm 0.002$ & $\mathbf{0.915 \pm 0.004}$ & $\mathbf{0.917 \pm 0.002}$ & $\mathbf{0.911 \pm 0.006}$ & $\mathbf{0.910 \pm 0.009}$ \\
\bottomrule
\end{tabular}}
\end{table}

The result is unambiguous: with one-time calibration of $\kappa$,
Adam-InitOnly yields test accuracy $0.91 \pm 0.005$ across
\emph{all} six learning rates, 5 seeds each, with zero divergences.
This mirrors and even strengthens the CIFAR-10
divergence-prevention result: across more than three orders of magnitude
of misspecification ($10^{-3}$ to $3.0$), test accuracy drops by
less than 1 percentage point relative to the optimal vanilla-Adam
configuration. The direction-matched probe achieves the same without
the sweep (final block; Figure~\ref{fig:rescue-headline}, bottom left):
$0.90$--$0.91$ across all six rates, zero
divergences, at the fixed $\kappa = 2$. The two parameter-free baselines (last block of
Table~\ref{tab:fashion-rescue}) again cover only part of the range:
Schedule-Free AdamW is the strongest method at $\eta \le 10^{-2}$ but
diverges for $\eta \ge 10^{-1}$, while Prodigy is robust from $10^{-2}$
to $3.0$ yet collapses to random-guess accuracy at $\eta = 10^{-3}$.
The two failure modes are complementary and opposite, and neither
matches the calibrated init probe's coverage of all six rates.

Adam-InitOnly outperforms the periodic controllers in this setting
because Adam's internal adaptation handles subsequent curvature
changes, whereas the periodic updates overshoot.
The Watchdog and Tracker modes were tuned on CIFAR-10/ResNet-18,
where progressive sharpening~\citep{cohen2021eos} is comparatively
gentle and the periodic adjustments tracked the curvature
dynamics usefully. On a small CNN with stronger progressive
sharpening, the same periodic adjustments overshoot: the Watchdog
shrinks $\eta$ down to its floor of $0.05\,\eta_{\text{init}}$, the
Tracker EMA pulls $\eta$ outside the stable range. Removing the
periodic component (Adam-InitOnly) lets Adam's own adaptive
$1/\sqrt{v_t}$ scaling handle in-training curvature changes, and
performs strictly better. This sharpens the recommendation: the
init probe is the primary mechanism. Periodic probing should be
treated as an optional, architecture-dependent extension.

This calibration is low-cost. The extra cost of fixing $\kappa$ to a new
architecture is a single divergence sweep like the one in
Appendix~\ref{sec:results-divergence} ($\sim 50$ Adam runs of $\le 1$
epoch each), or roughly one minute on the hardware of
Section~\ref{sec:setup}. Once
calibrated, Adam-InitOnly deploys with $<\!2\%$ wall-clock overhead per
training run.

To test whether the recipe extends beyond vision, we repeat the study on
the text benchmark: AG News with the small Transformer. The probe and
controllers are unchanged; results
over the same $6\times5$ grid are in Table~\ref{tab:agnews-rescue}.

\begin{table}[htbp]
\centering
\caption{AG News / small Transformer: uncalibrated $\kappa = 0.25$ versus
calibrated $\kappa = 0.05$. Mean$\pm$std over 5 seeds; best per column
within each block in bold. ``DIV.''\ = all
seeds diverged (loss $\to$ NaN); a bare $0.250$ is random-guess accuracy
(4 balanced classes),
i.e.\ the run was non-divergent but unproductive; ``$x/n$ div.''\ = partial
divergence with surviving-seed mean. The largest learning rate that still
trains above chance here is $\eta \approx 10^{-3}$, an order of magnitude
lower than on CIFAR-10, where vanilla Adam is still productive at $10^{-2}$;
the CIFAR default $\kappa = 0.25$ therefore caps above the productive range
and does not prevent divergence, exactly as on Fashion-MNIST. A
one-minute divergence sweep gives $\kappa = 0.05$, after which Adam-InitOnly
is robust across all six rates.}
\label{tab:agnews-rescue}
\resizebox{\textwidth}{!}{%
\begin{tabular}{lcccccc}
\toprule
$\eta_{\text{init}}$ & $10^{-3}$ & $10^{-2}$ & $10^{-1}$ & $0.3$ & $1.0$ & $3.0$ \\
\midrule
\multicolumn{7}{l}{\textit{Uncalibrated, $\kappa = 0.25$ (CIFAR-10 default)}} \\
Vanilla Adam       & $0.902 \pm 0.003$ & $0.250$ & DIV.\ & DIV.\ & DIV.\ & DIV.\ \\
Adam-InitOnly      & $\mathbf{0.905 \pm 0.003}$ & $0.250$ & $0.250$ & $0.250$ & $0.250$ & $0.250$ \\
Adam-Watchdog      & $0.904 \pm 0.002$ & $0.250$ & $0.250$ & DIV.\ & DIV.\ & DIV.\ \\
Adam-Tracker       & DIV.\ & DIV.\ & DIV.\ & DIV.\ & DIV.\ & DIV.\ \\
Schedule-Free AdamW & $0.903 \pm 0.003$ & $0.251$ & 0.250\,(3/5 div.) & DIV.\ & DIV.\ & DIV.\ \\
Prodigy            & $0.232 \pm 0.007$ & $\mathbf{0.880 \pm 0.002}$ & $\mathbf{0.906 \pm 0.002}$ & $\mathbf{0.908 \pm 0.002}$ & $0.408 \pm 0.195$ & $0.259 \pm 0.018$ \\
\midrule
\multicolumn{7}{l}{\textit{Calibrated, $\kappa = 0.05$ (one extra divergence sweep, $<\!1$ minute)}} \\
\textbf{Adam-InitOnly} & $\mathbf{0.905 \pm 0.003}$ & $\mathbf{0.909 \pm 0.002}$ & $\mathbf{0.909 \pm 0.002}$ & $\mathbf{0.909 \pm 0.002}$ & $\mathbf{0.909 \pm 0.002}$ & $\mathbf{0.909 \pm 0.002}$ \\
Adam-Watchdog      & $0.904 \pm 0.002$ & $0.250$ & $0.250$ & DIV.\ & DIV.\ & DIV.\ \\
Adam-Tracker       & DIV.\ & DIV.\ & DIV.\ & DIV.\ & DIV.\ & DIV.\ \\
\midrule
\multicolumn{7}{l}{\textit{Direction-matched probe, fixed $\kappa = 2$ (no calibration)}} \\
\textbf{Adam-InitOnly (dir.)} & $\mathbf{0.904 \pm 0.003}$ & $0.250$ & $0.250$ & $0.250$ & $0.250$ & $0.250$ \\
\bottomrule
\end{tabular}}
\end{table}

The pattern mirrors Fashion-MNIST closely. With the CIFAR default
$\kappa = 0.25$, no method is robust: vanilla Adam, Adam-InitOnly and
Adam-Watchdog all collapse to random-guess accuracy the moment
$\eta \ge 10^{-2}$, because the largest stable Adam learning rate on this
Transformer is only $\approx 3\times10^{-3}$ and the probe's safe-step
estimate at initialisation ($\bar\alpha_{\text{init}} \approx 0.031$) caps
$\eta$ at $0.25\,\bar\alpha_{\text{init}} \approx 8\times10^{-3}$, still
above the stable range. The two parameter-free baselines again cover only
part of the grid: Schedule-Free AdamW fails for $\eta \ge 10^{-2}$, and
Prodigy is robust only in a middle band ($10^{-2}$ to $0.3$) while failing
at both ends. After the divergence sweep sets
$\kappa = 0.05$, Adam-InitOnly recovers $0.905$--$0.909$ across all
six learning rates, more than three orders of magnitude of misspecification, with zero
divergences. The periodic controllers do not transfer even after
recalibration, confirming that the one-shot init probe, not the periodic
machinery, is the component that generalises across domains.

The direction-matched probe marks the limit of the calibration-free
recipe here (final block of Table~\ref{tab:agnews-rescue};
Figure~\ref{fig:rescue-headline}, bottom right). At the fixed
$\kappa = 2$ it still prevents every divergence, but on this
Transformer, whose productive range is unusually narrow
($\eta \approx 10^{-3}$), the uncalibrated cap lands in the
non-divergent-yet-unproductive band, and the runs sit at chance for
$\eta \ge 10^{-2}$. Where the productive range is that narrow, the fixed cap
is a divergence safeguard only; the calibration sweep moves the cap
into the productive range and restores accuracy across the grid.

Imagenette, the real-resolution control of
Section~\ref{sec:setup}, completes the transfer picture. Vanilla Adam
repeats the CIFAR-10 pattern exactly, competitive at $\eta = 10^{-3}$
and diverging in the first step from $\eta = 0.1$ on, and the fixed
$\kappa = 2$ again removes every divergence while staying within about
five points of the tuned optimum (Figure~\ref{fig:rescue-headline},
top right; exact numbers in Appendix~\ref{app:imagenette}). The divergence-prevention result
thus carries over from thumbnails to real images and to the unmodified
stem.

A sharper test is the case where the default learning rate is itself
unsafe. On every
benchmark above the Adam framework default $\eta = 10^{-3}$ lies in
the stable range, so the init cap never fires and training is
untouched. Its value appears only
where the default is genuinely too large for the curvature at
initialisation. Two well-known such regimes are normalisation-free networks
and warmup-free deep Transformers (Table~\ref{tab:default-unsafe}): a
ResNet-18 with all normalisation removed on CIFAR-10, and a deeper $12$-layer
variant of the AG News Transformer (versus the $4$-layer model used above)
trained without warmup. At the default
$\eta = 10^{-3}$ vanilla Adam \emph{diverges} (no-normalisation ResNet, all
seeds) or \emph{collapses to chance} (warmup-free Transformer); Adam-InitOnly,
with $\kappa$ recalibrated by the same sweep, caps the
rate into the stable range and recovers accuracy within a point or two of
the best tuned vanilla Adam, without the user knowing the safe rate in
advance. The caveat is that the CIFAR default $\kappa = 0.25$ does not
transfer to these higher-curvature settings (it still diverges). Together with the CIFAR-10
grid, these results support the divergence-prevention claim: the
one-shot cap prevents divergence where the default rate is genuinely
unstable and is a no-op otherwise.

\begin{table}[htbp]
\centering
\caption{Where the default learning rate is itself unsafe. At Adam's
framework default $\eta = 10^{-3}$, a normalisation-free ResNet and a
warmup-free deep Transformer diverge or fail to learn; the one-shot init
cap (with a per-architecture $\kappa$, set by a one-minute divergence sweep)
recovers accuracy within a point or two of the best tuned vanilla Adam,
whereas the default $\kappa = 0.25$ does not transfer. Mean over 4 seeds;
short protocol (ResNet 3 epochs, Transformer 600 steps).}
\label{tab:default-unsafe}
\resizebox{\textwidth}{!}{%
\begin{tabular}{lccc}
\toprule
Setting & vanilla, default $10^{-3}$ & vanilla, best tuned & Adam-InitOnly @ $10^{-3}$ \\
\midrule
ResNet-18, \emph{no normalisation} (CIFAR-10) & DIV.\ (4/4) & $0.537$ ($\eta{=}3{\times}10^{-4}$) & $0.535$ ($\kappa{=}0.05$) \\
Transformer, \emph{no warmup} (AG News) & $0.250$ (chance) & $0.828$ ($\eta{=}3{\times}10^{-4}$) & $0.810$ ($\kappa{=}0.005$) \\
\bottomrule
\end{tabular}}
\end{table}

Two further checks (the empirical divergence boundary that fixes
$\kappa$, and a 100-epoch cosine schedule on which the divergence-prevented runs
retain their full $\approx 93\%$ accuracy across the entire grid) are
deferred to
Appendices~\ref{sec:results-divergence} and \ref{sec:results-longrun}.

The divergence-prevention results above each used a per-architecture $\kappa$, set by a one-minute
divergence sweep, and this is their main practical drawback. It is, however, a
property of the \emph{probed direction}, not of the method, and the
direction-matched probe removes it. The plain-gradient probe reports
$\bar\alpha_{\text{init}}$ in a coordinate system Adam never steps in, so the
conversion factor that $\kappa$ must absorb depends on the per-architecture
gradient scale. Probing instead along Adam's first-step preconditioned
direction $-g/(|g|+\varepsilon) \approx -\operatorname{sign}(g)$ returns
$\bar\alpha_{\text{init}}$ in the geometry Adam actually moves in, and that
dependence disappears. Table~\ref{tab:universal-kappa} confirms it. A single fixed
$\kappa = 3$, chosen on four architectures and held out on five unseen
ones, caps safely across nine architectures (BatchNorm, GroupNorm and
normalisation-free nets; a plain MLP; an LSTM; CNNs and Transformers of
several scales) and two learning rates: $2$ divergences in $54$ runs,
against $14$ for the plain-gradient probe at a single $\kappa$ and $25$
for vanilla Adam, and \emph{zero} at the framework default. Both
failures occur on a plain MLP at the $100\times$-too-large rate, and a
sweep of the safety factor (Table~\ref{tab:kappa-universal-sweep})
removes them: $\kappa = 2$ and even $\kappa = 1$ give zero divergences
in all $54$ runs; $\kappa = 2$ does so at no accuracy cost, while the more
conservative $\kappa = 1$ trades about two points at the default rate
($0.704$ against $0.721$ for $\kappa = 3$). We therefore recommend
$\kappa = 2$ as the less intrusive of the two divergence-free settings.

The selection history bounds what this sweep alone can claim: $\kappa = 3$
was chosen on four architectures and confirmed on five held-out ones, but
$\kappa = 2$ is selected \emph{on} the full suite, so the nine
architectures cannot double as its held-out validation. We therefore
validate the fixed $\kappa = 2$ on three genuinely held-out
architectures, never used in any selection or tuning step of this
project: a small Vision Transformer ($2.7$M parameters, patch $4$) on
CIFAR-10, a two-layer GRU on AG News, and MobileNetV2 (stride-$1$ stem)
on CIFAR-10 (Table~\ref{tab:heldout-kappa2}). At the same fixed
$\kappa = 2$, the static and probationary variants give zero divergences
in all $36$ capped runs, while vanilla Adam at $\eta = 0.1$ diverges on
MobileNetV2 and the ViT ($3/3$ each) and survives at chance accuracy
on the GRU ($0.256$); capped accuracy matches the default-rate accuracy
on ViT and GRU and exceeds it on MobileNetV2. The init readings also
confirm the saturation discussion on unseen architectures: the ViT reads
$\bar\alpha = 4.1 \times 10^{-4}$, below the $\beta^{8}$ floor, so the
extended-backtracking init reading is what makes the cap correct there
(Table~\ref{tab:protocol}).

\begin{table}[htbp]
\centering
\caption{Genuine held-out validation of the fixed $\kappa = 2$: three
architectures never used in any selection or tuning step, at the suite
budget (3 epochs; 600 steps on AG News; 3 seeds; conventions as in
Table~\ref{tab:rescue}). Zero divergences in all $36$ capped runs.}
\label{tab:heldout-kappa2}
\resizebox{\textwidth}{!}{%
\begin{tabular}{lcccccc}
\toprule
 & \multicolumn{2}{c}{vanilla Adam} & \multicolumn{2}{c}{Adam-InitOnly (dir., $\kappa = 2$)} & \multicolumn{2}{c}{+ probation} \\
$\eta_{\text{init}}$ & $10^{-3}$ & $10^{-1}$ & $10^{-3}$ & $10^{-1}$ & $10^{-3}$ & $10^{-1}$ \\
\midrule
ViT small (CIFAR-10) & $0.503 \pm 0.008$ & DIV. & $0.499 \pm 0.026$ & $0.502 \pm 0.011$ & $0.511 \pm 0.013$ & $0.518 \pm 0.010$ \\
GRU (AG News) & $0.879 \pm 0.003$ & $0.256 \pm 0.005$ & $0.872 \pm 0.004$ & $0.875 \pm 0.019$ & $0.879 \pm 0.005$ & $0.894 \pm 0.015$ \\
MobileNetV2 (CIFAR-10) & $0.556 \pm 0.002$ & DIV. & $0.551 \pm 0.006$ & $0.629 \pm 0.014$ & $0.566 \pm 0.010$ & $0.629 \pm 0.010$ \\
\bottomrule
\end{tabular}}
\end{table}

\begin{figure}[htbp]
\centering
\includegraphics[width=0.92\linewidth]{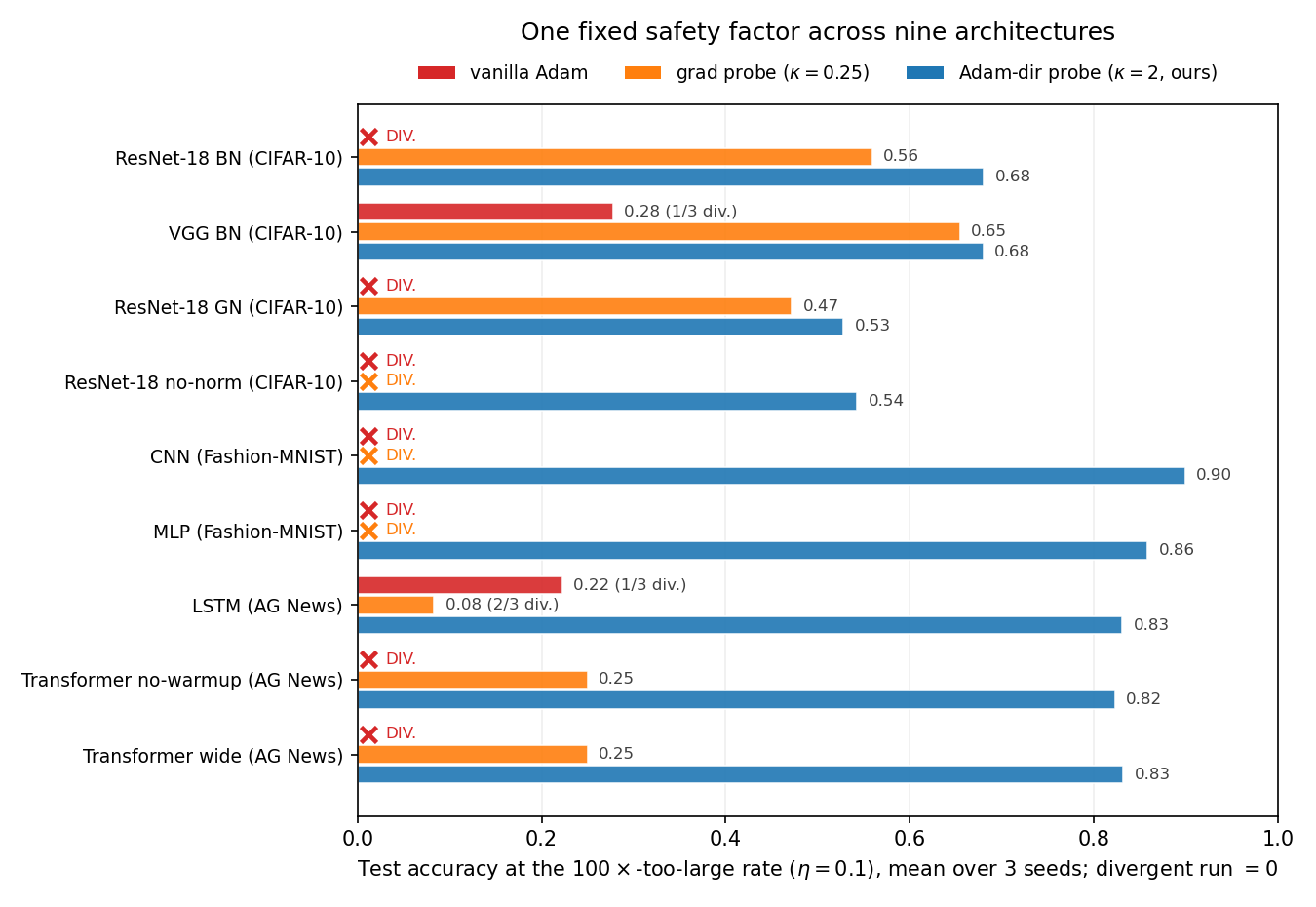}
\caption{The transfer result at a glance: per-architecture test accuracy at
the $100\times$-too-large learning rate ($\eta = 0.1$; mean over 3 seeds,
divergent run counted as $0$). With a single fixed safety factor, the
direction-matched probe ($\kappa = 2$, blue) trains on all nine
architectures and is the most accurate variant on each, whereas vanilla
Adam (red) diverges on seven of the nine and the plain-gradient probe at
one fixed $\kappa = 0.25$ (orange) diverges completely on three, partially
on a fourth, and degrades on others.
Exact values in Table~\ref{tab:kappa-perarch}.}
\label{fig:kappa-perarch}
\end{figure}

Accuracy tells
the same story (Table~\ref{tab:universal-kappa}, lower block): counting
a divergent run as zero, the direction-matched probe reaches a mean of
$0.72/0.63$ at the default/$100\times$ rate, against $0.55/0.06$ for
vanilla Adam and $0.51/0.25$ for the plain-gradient probe. The
default-rate gap deserves to be made explicit rather than left inside
the aggregate, because it shows the cap \emph{binding} at the framework
default: with the suite's extended-backtracking init reading
(Table~\ref{tab:protocol}), the cap engages below $10^{-3}$ in $14$ of
$54$ probe runs, and on both no-warmup AG News Transformers this is what
prevents divergence during training -- vanilla Adam at the default rate survives but sits
at chance accuracy ($0.25$), while the capped runs reach $0.84$. The
framework default is not ``already safe'' on these architectures, so the
intervention is consistent with the no-op property, which is scoped to
safe rates; on the architectures where $10^{-3}$ \emph{is} safe, the cap
does not bind and the probe leaves state and first update bit-identical
(Appendix~\ref{app:noop}). Because the
nine architectures have different chance levels and task difficulties,
Figure~\ref{fig:kappa-perarch} breaks the discriminating $100\times$
rate down per architecture: at $\kappa = 2$ the direction-matched probe
is the most accurate variant on each and diverges on none (exact numbers
in Appendix~\ref{app:kappa-perarch}). The full learning-rate grids
extend this beyond the two spot rates: at the same fixed $\kappa = 2$
the direction-matched probe diverges nowhere up to $\eta = 3.0$ on
all four benchmarks, with full accuracy recovery on CIFAR-10 and
Fashion-MNIST and, on Imagenette, recovery to within about five points of
the tuned optimum (Appendix~\ref{app:imagenette}) and the AG News limit discussed above
(final blocks of Tables~\ref{tab:rescue}, \ref{tab:fashion-rescue}
and~\ref{tab:agnews-rescue}). The same recipe also transfers to AdamW
unchanged (AdamW block of Table~\ref{tab:rescue}). The per-architecture $\kappa$ of the raw-gradient probe is thus a
limitation of that probe rather than of the safeguard itself.

\begin{table}[htbp]
\centering
\caption{A single safety factor across the nine-architecture suite.
Divergent runs and
mean test accuracy over nine architectures $\times$ \{default $\eta = 10^{-3}$,
$100\times$ too large\} $\times$ 3 seeds, under the suite's deliberately
short budget (3 epochs; 600 steps on AG News; Table~\ref{tab:protocol}) --
accuracies are not comparable to the 10-epoch grids of
Table~\ref{tab:rescue}. Probing along Adam's preconditioned
direction lets one fixed $\kappa = 3$ cap safely almost everywhere (with zero
divergences at the default rate), whereas the plain-gradient probe needs a
per-architecture $\kappa$. Accuracy counts a divergent run as $0$, so it
rewards both not diverging and still learning; the direction-matched probe is
best on both rows. Both direction-matched failures occur on a plain MLP at
the $100\times$ rate.}
\label{tab:universal-kappa}
\resizebox{\textwidth}{!}{%
\begin{tabular}{lccc}
\toprule
 & vanilla Adam & plain-grad probe ($\kappa = 0.25$) & Adam-direction probe ($\kappa = 3$) \\
\midrule
\multicolumn{4}{l}{\textit{Divergent runs}} \\
at default $\eta = 10^{-3}$ (of 27) & $2$ & $3$ & $\mathbf{0}$ \\
at $100\times$ too large (of 27)    & $23$ & $11$ & $\mathbf{2}$ \\
total (of 54)                       & $25$ & $14$ & $\mathbf{2}$ \\
\midrule
\multicolumn{4}{l}{\textit{Mean test accuracy (divergent run $=0$)}} \\
at default $\eta = 10^{-3}$          & $0.552$ & $0.507$ & $\mathbf{0.721}$ \\
at $100\times$ too large            & $0.056$ & $0.252$ & $\mathbf{0.631}$ \\
\bottomrule
\end{tabular}}
\end{table}

\begin{table}[htbp]
\centering
\caption{Sweeping the safety factor of the Adam-direction probe over the same
nine architectures $\times$ two rates $\times$ three seeds (same short
budget as Table~\ref{tab:universal-kappa}). A more conservative
$\kappa$ removes the last divergences: $\kappa = 2$ and $\kappa = 1$ give
zero divergences in all $54$ runs. At $\kappa = 2$ this costs no accuracy --
it matches or beats $\kappa = 3$ at both rates ($0.729$ vs.\ $0.721$
at the default, $0.742$ vs.\ $0.631$ at $100\times$, where $\kappa = 3$ also
diverges twice). We recommend $\kappa = 2$.}
\label{tab:kappa-universal-sweep}
\begin{tabular}{lccc}
\toprule
Adam-direction probe & $\kappa = 1$ & $\kappa = 2$ & $\kappa = 3$ \\
\midrule
divergences at default $10^{-3}$ (of 27) & $\mathbf{0}$ & $\mathbf{0}$ & $0$ \\
divergences at $100\times$ (of 27)       & $\mathbf{0}$ & $\mathbf{0}$ & $2$ \\
total divergences (of 54)                & $\mathbf{0}$ & $\mathbf{0}$ & $2$ \\
\midrule
mean acc at default $10^{-3}$            & $0.704$ & $\mathbf{0.729}$ & $0.721$ \\
mean acc at $100\times$                  & $0.732$ & $\mathbf{0.742}$ & $0.631$ \\
\bottomrule
\end{tabular}
\end{table}

\phantomsection\label{sec:results-lm}
We next evaluate the method on language-model pretraining, where a
divergent run is substantially more expensive. We test this directly, on
GPT-2~\citep{radford2019gpt2}
architectures trained from scratch on WikiText-103~\citep{merity2017wikitext}
(GPT-2 BPE
tokenisation, $118$M training tokens): \emph{small} ($124$M parameters)
and \emph{medium} ($355$M), sequence length $512$, $8192$ tokens per
step, AdamW without warmup and without gradient clipping, a fixed
budget of $10^{4}$ steps ($\approx 82$M tokens), the learning-rate grid
and divergence criterion of the vision experiments, and validation
perplexity after the budget as the quality metric (two seeds
throughout; Table~\ref{tab:lm-rescue}).

The static init cap fails here, and not marginally. Vanilla AdamW
diverges within the first ten steps from $\eta = 0.04$ on, while the
init probe reads $\bar\alpha_{\text{Adam}} = 0.0625$, a robust value
($56$ of $60$ probe batches over three seeds agree to the rung), so the
divergence boundary lies \emph{below} the reading itself and the fixed
$\kappa = 2$ caps into divergent territory. Nor is a smaller constant
the answer: the sweep-calibrated $\kappa = 0.5$, which caps at the
conservative end $\beta\bar\alpha$ of the backtracking band, still
diverges on the full budget at every rate above the cap, at steps ranging
from $6$ to $5130$ across models, rates and seeds. The late failures are
what make a short check misleading: a run still healthy after $10^{3}$
steps can collapse thousands of steps later. A size ladder
within the same setup ($16$M, $45$M, $124$M, $355$M) shows the safety
margin fluctuating between $0.5\,\bar\alpha$ and $2\,\bar\alpha$ with
no clean size trend, against $\ge 2\,\bar\alpha$ on every classifier of
Table~\ref{tab:universal-kappa}: the thin margin is a property of the
decoder-LM training setup, not of parameter count.

\begin{figure}[htbp]
\centering
\includegraphics[width=0.78\linewidth]{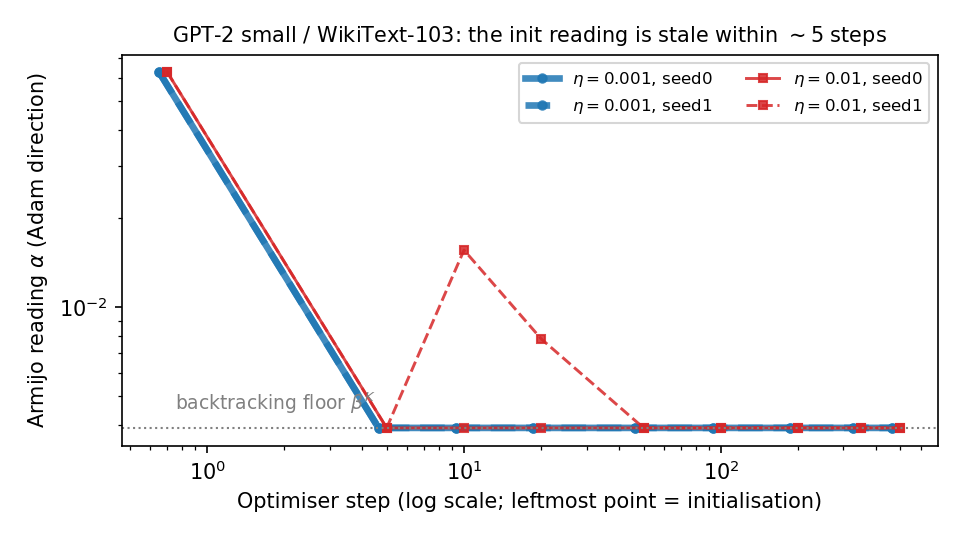}
\caption{Why the static init cap fails at language-model scale: the
direction-matched Armijo reading collapses from $0.0625$ at
initialisation to the backtracking floor $\beta^{K}$ within about five
optimiser steps of \emph{stable} GPT-2-small training (both stable
learning rates, both seeds; the transient recovery of one
$\eta = 0.01$ seed follows a catapult-style loss spike). A cap frozen
at step $0$ therefore overestimates the safe rate by a factor
${\ge}16$; the probationary window tracks the collapse and lands at
$\kappa\beta^{K} \approx 0.008$.}
\label{fig:lm-sharpening}
\end{figure}

Repeated probes identify the mechanism: the accepted step decreases by
at least a factor of 16 within the first five optimiser steps
(Figure~\ref{fig:lm-sharpening}). Re-running the probe during stable
training, the reading collapses from $0.0625$ at initialisation to the
backtracking floor $\beta^{K} = 0.0039$ within about five optimiser
steps, at both stable rates and in every seed: the loss surface along
the probe direction sharpens by a factor ${\ge}16$ immediately after the
first updates, so a step-$0$ reading is stale when the cap freezes it.
This is progressive sharpening at initialisation, the same phenomenon
that motivates warmup at LM scale~\citep{kalra2024warmup}, here read at
five forward passes per point. It also explains, retroactively, why the
fixed $\kappa$ held on the classifier suite: there the initial reading
already sits at the floor $\beta^{K}$ (Table~\ref{tab:probe-variance}),
so the initialisation value is already at the backtracking floor, and
the same mismatch does not arise.

The probationary window of Section~\ref{sec:method} addresses this
mismatch without changing $\kappa$. Re-probing during the first $50$
steps updates the cap to $\kappa\beta^{K} \approx 0.008$ within at
most eight steps, and
the resulting runs neither diverge nor stall: zero divergences in all
$24$ grid runs per model size, with validation perplexity inside the
band of the tuned default across the entire grid up to $\eta = 3.0$,
on both model sizes. The warmup baseline sharpens the comparison.
Linear warmup over the first $10\%$ of the budget delivers the best
default-rate quality on the small model (perplexity $466 \pm 101$ over
four seeds, against roughly $890$--$1030$ without warmup; run-to-run
spread is large in this regime) but diverges during the ramp for every
$\eta \ge 0.1$: warmup can improve optimisation quality but does not
prevent divergence when its target rate is too large. Guarding
the ramp with the probe (the probe caps the warmup \emph{target},
probationary window included) keeps warmup's default-rate quality
within the same spread ($600 \pm 64$) and never diverges either. At capped rates both probe variants inherit the
no-warmup plateau of this regime rather than the tuned-default quality,
consistent with the scoped claim throughout this paper: the cap
prevents divergence but does not restore accuracy. At language-model scale, the
measurement must be repeated during the first 50 steps rather than
taken only at initialisation.

\begin{table}[htbp]
\centering
\caption{Language-model pretraining divergence prevention (WikiText-103, AdamW, no
warmup unless stated, no gradient clipping, $10^{4}$ steps
$\approx 82$M tokens). Validation perplexity after the fixed budget
(lower is better), mean$\pm$std over 2 seeds; ``DIV.''\ = all seeds
diverged (loss above $5\times$ its initial value or non-finite);
``$x/n$ div.''\ = partial divergence with surviving-seed mean. The two
warmup rows at $\eta = 10^{-3}$ on the small model average $4$ seeds.
Static caps fail (the $\kappa = 0.5$ rows diverge at steps $6$ to
$5130$, the latest of them long after a short check would have passed); the probationary probe and the probe-guarded warmup
never diverge anywhere on the grid.}
\label{tab:lm-rescue}
\resizebox{\textwidth}{!}{%
\begin{tabular}{lcccccc}
\toprule
$\eta_{\text{init}}$ & $10^{-3}$ & $10^{-2}$ & $10^{-1}$ & $0.3$ & $1.0$ & $3.0$ \\
\midrule
\multicolumn{7}{l}{\textit{GPT-2 small ($124$M)}} \\
Vanilla AdamW & $959.2 \pm 66.1$ & $1026.4 \pm 41.1$ & DIV.\ & DIV.\ & DIV.\ & DIV.\ \\
AdamW + linear warmup & $466.3 \pm 100.9$ & $1033.3 \pm 79.4$ & DIV.\ & DIV.\ & DIV.\ & DIV.\ \\
InitOnly (dir., $\kappa = 2$) & $871.4 \pm 55.5$ & $1074.0 \pm 20.8$ & DIV.\ & DIV.\ & DIV.\ & DIV.\ \\
InitOnly (dir., sweep $\kappa = 0.5$) & $921.1 \pm 3.3$ & $1023.6 \pm 0.1$ & DIV.\ & DIV.\ & DIV.\ & DIV.\ \\
\textbf{Probation (dir., $\kappa = 2$)} & $889.1 \pm 73.4$ & $998.6 \pm 82.2$ & $978.5 \pm 130.7$ & $1082.9 \pm 29.4$ & $910.8 \pm 5.9$ & $1251.6 \pm 261.0$ \\
\textbf{Guarded warmup} & $599.6 \pm 63.5$ & $1037.9 \pm 51.7$ & $1093.0 \pm 114.3$ & $1014.7 \pm 139.8$ & $942.4 \pm 75.5$ & $1106.4 \pm 6.3$ \\
\midrule
\multicolumn{7}{l}{\textit{GPT-2 medium ($355$M)}} \\
Vanilla AdamW & $1041.2 \pm 15.9$ & 1769.9\,(1/2 div.) & DIV.\ & DIV.\ & DIV.\ & DIV.\ \\
AdamW + linear warmup & $1219.5 \pm 120.3$ & $1158.6 \pm 89.3$ & DIV.\ & DIV.\ & DIV.\ & DIV.\ \\
InitOnly (dir., $\kappa = 2$) & $1181.5 \pm 1.8$ & $1129.7 \pm 94.4$ & DIV.\ & DIV.\ & DIV.\ & DIV.\ \\
InitOnly (dir., sweep $\kappa = 0.5$) & $1169.9 \pm 11.1$ & $1300.2 \pm 122.7$ & DIV.\ & DIV.\ & DIV.\ & DIV.\ \\
\textbf{Probation (dir., $\kappa = 2$)} & $1273.5 \pm 243.4$ & $929.1 \pm 20.0$ & $979.0 \pm 77.9$ & $992.3 \pm 52.5$ & $1011.1 \pm 27.4$ & $1050.4 \pm 0.9$ \\
\textbf{Guarded warmup} & $1099.0 \pm 43.4$ & $1163.8 \pm 16.4$ & $1091.6 \pm 47.9$ & $1064.7 \pm 45.2$ & $1296.8 \pm 205.5$ & $1240.8 \pm 22.2$ \\
\bottomrule
\end{tabular}}
\end{table}

\phantomsection\label{sec:results-cost}
Finally we turn to cost.
Wall-clock overhead in our setting is dominated by the explicit Hessian
probes used to measure the $\alpha$--$\lambda_1$ correlation above; the probes are an instrumentation
artefact and are not part of the deployed controller. We report the two
sets separately. With Hessian probes turned on, vanilla Adam takes 73.1
seconds for ten CIFAR-10 epochs, Adam-Tracker takes 76.1 seconds (+4\%),
and Adam-Watchdog takes 80.4 seconds (+10\%); the difference is the
extra forward and forward-backward passes the periodic Armijo probe
performs. With Hessian probes turned off (the deployed
configuration), the cost depends on how often the controller probes.
The three deployments cost differently. Adam-InitOnly probes once, at
initialisation: one extra forward-and-backward pass plus a handful of
forward-only backtracks, some $50$\,ms here, which against a ten-epoch
run is a tenth of a percent and within measurement noise of vanilla Adam;
the ``about $1\%$'' we quote elsewhere is the conservative figure for the
deployed protocol, whose probationary window adds up to nine further
probes over the first fifty steps. The periodic controllers cost more. In a controlled interleaved
benchmark (3 seeds, single session, GPU warmed up) over ten CIFAR-10
epochs, vanilla Adam takes $56.9$ s; Adam-Tracker adds $+2.7\%$ and
Adam-Watchdog $+5.3\%$. These figures use forward-only backtracking;
recomputing a gradient at every backtracking trial (the naive
implementation) roughly doubles the periodic overhead, with Adam-Watchdog
rising from $+5.3\%$ to $+12.0\%$, which is why the probe evaluates
the trial loss without a backward pass.

Several limitations bound these claims. The first, the per-architecture
$\kappa$ of the plain-gradient probe, was addressed above: it is a
property of the probed direction, and the direction-matched probe
removes it.

A second limitation, related to the first, is the fragility of the
periodic-probe controllers under rapid progressive sharpening: on the
small CNN their pre-tuned thresholds and EMA constants overshoot the
true sharpness boundary, and even after $\kappa$ is recalibrated both
underperform Adam-InitOnly. We accordingly keep Watchdog and Tracker as
opt-in features for users who want continuous in-training adaptation.

To quantify this, we compared the periodic Tracker
across regimes at the framework default $\eta = 10^{-3}$
(Table~\ref{tab:periodic}). On well-conditioned real tasks its step-size
growth overshoots and \emph{hurts}: it loses several accuracy points on
CIFAR-10/ResNet-18 and diverges on the AG-News Transformer
($5/5$ seeds), at \emph{every} probe interval $K$ we tried, so
the failure is intrinsic to raising the rate on a smooth landscape rather
than a matter of tuning $K$. On a plateau-dominated toy (a $2$-$2$ XOR
network, 100 seeds) the Tracker does help, with $66\%$ of seeds solved versus
$52\%$ at the default, but a fixed higher learning rate ($\eta = 0.1$)
already solves $61\%$, so most of the gain is that of an effectively
larger step rather than of online adaptivity. We also note that where a
task is hard for representational rather than step-size reasons (sharp
discontinuities, high-frequency targets), the default rate is already in
the stable range and the init cap does not fire; the probe is, by design,
silent there. These observations bound the claims to the one-shot init
probe, on which the paper rests.

\begin{table}[htbp]
\centering
\caption{Periodic in-training probing across regimes, at the default
$\eta = 10^{-3}$. The Tracker's step-size growth helps only on a
plateau-dominated toy, and there no more than a fixed higher learning
rate would; on well-conditioned real tasks it degrades or diverges
at every probe interval. Mean over seeds (5 for the real tasks, 100 for
XOR); on the XOR toy the Tracker uses its best probe interval ($K=20$), on
the real tasks the default $K=50$.}
\label{tab:periodic}
\begin{tabular}{lccc}
\toprule
Regime (default $\eta = 10^{-3}$) & vanilla Adam & Watchdog & Tracker \\
\midrule
Smooth: CIFAR-10 / ResNet-18 (test acc) & $0.834$ & $0.808$ & $0.803$ \\
Smooth: AG News / Transformer (test acc) & $0.902$ & $0.904$ & DIV.\ \\
Plateau toy: $2$-$2$ XOR (\% solved) & $52$ & $52$ & $66$ \\
\bottomrule
\end{tabular}
\end{table}

In the same spirit we tested whether the probe could serve as an
in-training \emph{early-warning} signal for instability, rather than a
controller. In our tests it did not provide a reliable early-warning signal
for runtime divergence, for two reasons we observed across mini-batch (LSTM
language modelling on War \& Peace, CNNs) and full-batch Edge-of-Stability
regimes: the reading does not discriminate (local curvature rises during
\emph{healthy} Edge-of-Stability training as much as before a blow-up),
and the divergences we could induce were too abrupt to leave actionable lead
time. On the evidence we have, the probe's value is therefore at
initialisation (a curvature sensor and a safety cap), not as a runtime
monitor; whether a refined variant could recover a usable signal we leave
open.

Finally, the empirical scope of this paper is deliberately narrow. We
report controlled experiments on three image-classification benchmarks
(CIFAR-10 with a ResNet-18, Fashion-MNIST with a small two-conv CNN,
and Imagenette with a standard ResNet-18 at $128$px)
one text-classification benchmark (AG News with a small from-scratch
Transformer), and language-model pretraining up to GPT-2 medium ($355$M
parameters) on WikiText-103 (Section~\ref{sec:results-lm}), studied
throughout with full learning-rate grids and multiple seeds per cell
rather than with single runs of larger models; we do not
test billion-parameter language models, self-supervised vision
pretraining, or ImageNet scale, the regimes where a diverged run is
even more expensive and a safeguard would matter most. The
empirical link between the line-search step and the top Hessian
eigenvalue is established for three architectures, not claimed
universal. The Discussion below sketches the experiments that would,
in our view, justify a stronger general claim.

\section{Discussion}

We have shown that classical Armijo line search, used as a probe rather
than as the optimiser, robustifies Adam against a specific but common
failure mode: an over-large initial learning rate. Its soundness rests
on a clean, empirically reproducible link between the
line-search step and the top Hessian eigenvalue.

The probe acts as an initialisation-time validation step. If the
user-specified learning rate is below the measured cap, training is
unchanged. Otherwise, the learning rate is reduced before the first
update.

The mechanism is optimiser-agnostic in principle, and a first transfer
check supports this: on CIFAR-10/ResNet-18,
AdamW~\citep{loshchilov2019adamw} diverges for every $\eta \ge 0.1$
exactly like Adam, and the direction-matched probe at the same fixed
$\kappa = 2$ removes every divergence while holding test accuracy at
$0.82$--$0.83$ across the full grid (AdamW block of
Table~\ref{tab:rescue}). Whether
the same holds for optimisers with a different first-step geometry,
such as RMSProp~\citep{tieleman2012rmsprop}, we leave open.

\phantomsection\label{sec:future}
The main remaining direction is validation at larger scale.
Section~\ref{sec:results-lm} takes the first step, GPT-2 pretraining up
to $355$M parameters, and the regime indeed behaves differently: the
divergence boundary drops below the initialisation reading, the static
cap fails, and only the probationary window restores the recipe.
Whether that pattern persists for billion-parameter language models,
self-supervised pretraining, or ImageNet-scale training remains
untested and we leave it to future work, together with the question of whether
the probe-as-controller idea is best combined with, rather than compared
against, the parameter-free optimisers
(D-Adaptation~\citep{defazio2023dadaptation} and
Mechanic~\citep{cutkosky2023mechanic}, which we discuss but do not run,
would belong in that comparison).

\section{Conclusion}

A single Armijo backtracking line search (five forward evaluations and
one backward pass) is a low-cost, Hessian-free sensor for local
sharpness: the step empirically inverts the \emph{directional} curvature
$q = g^\top H g/\|g\|^2$ within the derived backtracking band, and
$1/\alpha$ tracks $\lambda_1$ at Pearson $-0.91$ to $-0.95$ across three
architectures, $-0.60$ to $-0.70$ after removing the shared training
trend (Table~\ref{tab:correlation-detrended}): the textbook relation
$\alpha^\star \approx 1/\lambda_1$, recovered for the price of a few
forward passes rather than Lanczos iteration. The
practical pay-off is a one-shot learning-rate cap that keeps Adam within
a few points of its tuned accuracy across $\eta \in [10^{-3}, 3.0]$ on
CIFAR-10, Fashion-MNIST, Imagenette and AG News, where vanilla Adam diverges from
$\eta = 0.1$ on. Probing along Adam's first-step preconditioned direction
makes the multiplier transferable across the tested architectures. The
fixed protocol uses $\kappa = 2$ and the smallest reading obtained
during initialisation and the first 50 optimiser steps. It prevents
divergence across the full learning-rate grids and at GPT-2
pretraining scale. In the broader architecture study, one plain-MLP
setting remains sensitive to the initial batch order. The init-probe
cap transfers to AdamW unchanged, and the protocol incurs approximately
one percent overhead.
The method is intended to prevent divergence, not to guarantee
accuracy: recovery is full on the classifier benchmarks except the
narrow-productive-range AG News Transformer, while at GPT-2
pretraining scale, where the loss surface sharpens within the first
five optimiser steps and every cap frozen at initialisation fails, the
capped runs hold the quality of the no-warmup default. We hope the
low-cost curvature reading, more than the safeguard itself, encourages
the use of line-search quantities as loss-landscape diagnostics.

\FloatBarrier 

\appendix
\section{The negative result on the Golden-Section preselection}
\label{app:gs-failure}

Before using the line search as a measurement, we evaluated it as an
optimiser: a bounded golden-section preselection followed by Armijo
backtracking (\texttt{gs+armijo} in the released code, where the
line-search optimiser is named GGAO). It fails, and the failure is
instructive. The bounded
golden-section line search on $[0, 1]$ consistently selects very small
$\alpha$ ($\bar{\alpha}
\approx 0.05$ throughout training) and drives the trajectory into
regions of high $\lambda_1$ ($\bar\lambda_1 \approx 800$). Pure Armijo
on the same architecture and initialisation saturates at $\alpha = 1$
and stays in low-$\lambda_1$ regions ($\bar\lambda_1 \approx 200$). The
$\alpha$-$\lambda_1$ correlation, which is $-0.91 \pm 0.01$ for pure
Armijo on CIFAR-10/ResNet-18, drops to a non-significant $-0.32$ for GS+Armijo. The
golden-section step finds spurious early local minima of the noisy
mini-batch loss along the descent direction, where pure backtracking
correctly accepts the largest step that satisfies sufficient decrease.
We report this negative result here as a warning against bounded
non-monotone line searches on stochastic deep-learning losses.

\section{Sensitivity to the safety factor \texorpdfstring{$\kappa$}{kappa}}
\label{app:kappa-sensitivity}

The divergence-prevention mechanism depends on a single scalar, the safety factor $\kappa$, which
caps the user learning rate at $\kappa\,\bar\alpha_{\text{init}}$. To
quantify how forgiving this choice is, we sweep
$\kappa \in \{0.05, 0.1, 0.25, 0.5, 1.0\}$ for Adam-InitOnly on
CIFAR-10/ResNet-18 at three representative learning rates (the optimal
$10^{-3}$, and two values at which vanilla Adam diverges), 3 seeds each.
Table~\ref{tab:kappa-sens} reports the result.

\begin{table}[htbp]
\centering
\caption{Sensitivity of Adam-InitOnly to the safety factor $\kappa$ on
CIFAR-10/ResNet-18 (10 epochs, mean$\pm$std over 3 seeds). ``DIV.''\
= all seeds diverged. Any $\kappa \le 0.25$ preserves divergence prevention across
the whole range; only $\kappa \ge 0.5$ lets divergence back in at large
$\eta$. The procedure is thus insensitive to the exact value of $\kappa$
within a broad band, so the one-minute calibration sweep need not be
precise.}
\label{tab:kappa-sens}
\begin{tabular}{lccc}
\toprule
$\kappa$ & $\eta=10^{-3}$ & $\eta=10^{-1}$ & $\eta=1.0$ \\
\midrule
$0.05$ & $0.842 \pm 0.007$ & $\mathbf{0.824 \pm 0.016}$ & $\mathbf{0.829 \pm 0.019}$ \\
$0.1$  & $0.813 \pm 0.026$ & $0.812 \pm 0.012$ & $0.813 \pm 0.009$ \\
$0.25$ & $0.824 \pm 0.012$ & $0.794 \pm 0.011$ & $0.801 \pm 0.020$ \\
$0.5$  & $0.836 \pm 0.019$ & $0.735 \pm 0.033$ & $0.725 \pm 0.044$ \\
$1.0$  & $0.829 \pm 0.021$ & DIV.\ & DIV.\ \\
\bottomrule
\end{tabular}
\end{table}

At the optimal $\eta = 10^{-3}$ the cap rarely binds, so accuracy is
near-constant across $\kappa$. At the misspecified rates the trend is
monotone: smaller $\kappa$ caps the learning rate more aggressively and
preserves accuracy, whereas $\kappa = 1.0$ leaves the step essentially
uncapped and Adam diverges exactly as in the vanilla baseline. The
default $\kappa = 0.25$ used for CIFAR-10 sits comfortably inside the
safe band; divergence prevention degrades gracefully rather than catastrophically
as $\kappa$ grows, which is why a coarse one-minute calibration suffices.

\section{Cross-batch variance of the initialisation probe}
\label{app:probe-variance}

The reading of $\bar\alpha_{\text{init}}$ takes place on one mini-batch,
and therefore a natural question arises about how sensitive the reading and hence
the cap $\kappa\,\bar\alpha_{\text{init}}$ is to which batch is drawn.
This is exactly what is measured in Table~\ref{tab:probe-variance}, where for
each benchmark and three initialisation seeds the model is frozen at initialisation,
and the probe is run on $50$ different training mini-batches, in train mode, with the deployed
constants ($c = 10^{-4}$, $\beta = 0.5$, $\alpha_{\max} = 1$, $K = 8$), in both probe directions.
The runs were done with the unguarded probe of Appendix~\ref{app:noop}, and since at $\eta = 10^{-3}$
the cap is never active for any $\kappa$ in the experiment (the raw-gradient reading is $\bar\alpha = 0.0625$,
so even $\kappa = 0.05$ caps at $3.1 \times 10^{-3}$), with state restoration
they would look exactly alike; the variability between them represents the trajectory-level
noise of the probe, and not an effect of $\kappa$. Since $\alpha$ is discretised as powers of $\beta$,
the spread is measured in backtracking rungs.

The raw-gradient reading moves by at most one rung on eleven of the
twelve model--seed combinations and by two rungs in the worst case (one
AG News batch in $150$), so the implied cap varies across batches by at
most a factor $4$ once and a factor $2$ otherwise. This is within the
resolution that the backtracking band already concedes, and it is
negligible against the three decades of misspecification the safeguard
covers. The
Adam-direction probe at initialisation exhausts the backtracking cap
$K = 8$ of Algorithm~\ref{alg:init-probe} on all four benchmarks
($\alpha = \beta^8$ on almost every batch; Fashion-MNIST has a small number
of $\beta^7$ acceptances): at initialisation the curvature along
$-\operatorname{sign}(g)$ is large enough that only the one-sided bound
of Section~\ref{sec:results-rescue} applies, so the returned floor
value is batch-independent by construction and the fixed-$\kappa$
recipe inherits no single-batch noise at this point. The cap it
implies, $\kappa\,\beta^8 \approx 0.008$ at $\kappa = 2$, is the
effective rate behind the direction-matched rows of the divergence-prevention result tables:
near-optimal on the three image benchmarks, and inside the
stable-yet-unproductive band on AG News
(Section~\ref{sec:results-fashion}).

\begin{table}[htbp]
\centering
\caption{Cross-batch variance of the one-shot probe reading at a frozen
initialisation: median, min and max of $\alpha$ over $50$ training
batches, pooled over $3$ initialisation seeds (min/max are the extremes
over all $150$ readings). Rung span is the worst-case per-seed spread
in backtracking rungs ($\log_{1/\beta}$ units); the last column is the
share of readings within one rung of the per-seed median.}
\label{tab:probe-variance}
\resizebox{\textwidth}{!}{%
\begin{tabular}{llccccc}
\toprule
Benchmark & probe direction & median $\alpha$ & min & max & rung span & within $1$ rung \\
\midrule
CIFAR-10/ResNet-18 & raw gradient & $0.0625$ & $0.0625$ & $0.125$ & $1$ & $100\%$ \\
CIFAR-10/ResNet-18 & Adam-direction & $0.00391$ & $0.00391$ & $0.00391$ & $0$ & $100\%$ \\
Fashion-MNIST/CNN & raw gradient & $1$ & $0.5$ & $1$ & $1$ & $100\%$ \\
Fashion-MNIST/CNN & Adam-direction & $0.00391$ & $0.00391$ & $0.00781$ & $1$ & $100\%$ \\
AG News/Transformer & raw gradient & $0.0312$ & $0.0312$ & $0.125$ & $2$ & $99\%$ \\
AG News/Transformer & Adam-direction & $0.00391$ & $0.00391$ & $0.00391$ & $0$ & $100\%$ \\
Imagenette/ResNet-18 & raw gradient & $0.0625$ & $0.0312$ & $0.0625$ & $1$ & $100\%$ \\
Imagenette/ResNet-18 & Adam-direction & $0.00391$ & $0.00391$ & $0.00391$ & $0$ & $100\%$ \\
\bottomrule
\end{tabular}}
\end{table}

\section{Adam's divergence threshold and the safety factor \texorpdfstring{$\kappa$}{kappa}}
\label{sec:results-divergence}

The choice of $\kappa$ is made once, from the safety sweep of
Table~\ref{tab:kappa-universal-sweep} over the nine-architecture suite
($\kappa = 2$ for the direction-matched probe, zero divergences there),
and is then held fixed; its transfer is confirmed on the full
learning-rate grids, at GPT-2 scale, and on the three genuinely held-out
architectures of Table~\ref{tab:heldout-kappa2}. As a complementary check we here locate the \emph{absolute}
divergence boundary of Adam and confirm it sits close to the line-search
scale. For this single measurement we anchor to a low-variance estimate
$\bar{\alpha}$ obtained by averaging the Armijo step over a short (200-step)
warmup, rather than the raw init value, which is noisier and can saturate at
$\alpha = 1$ on low-curvature initialisations; this gives
$\bar{\alpha}_{\text{warmup}} = 0.0399$ on CIFAR-10/ResNet-18 (the finer of
two sweeps; a coarser one anchored at $0.0393$). We then trained Adam at
$\eta = m \cdot \bar{\alpha}$ for $m \in \{0.25, 0.4, 0.5, 0.6, 0.7,
0.8, 0.9, 1.0, 1.5, 2.0, 5.0\}$. Table~\ref{tab:kappa} summarises the
result.

\begin{table}[htbp]
\centering
\caption{Empirical determination of the absolute divergence boundary,
expressed as a multiple of the warmup-averaged Armijo anchor
$\bar{\alpha}_{\text{warmup}} = 0.0399$; note this anchor is measured over a
$200$-step warmup and is not the initialisation reading
$\bar\alpha_{\text{init}}$ of Table~\ref{tab:probe-variance}. Adam runs of
$400$ optimiser steps on CIFAR-10/ResNet-18.}
\label{tab:kappa}
\begin{tabular}{ccc}
\toprule
$m = \eta / \bar{\alpha}$ & $\eta$ & Outcome \\
\midrule
$0.25, 0.4, 0.5$  & $0.0099, 0.0160, 0.0200$ & all stable \\
$0.6$             & $0.0240$                  & diverged \\
$0.7$             & $0.0280$                  & stable (boundary jitter) \\
$0.8, 0.9, 1.0, 1.5, 2.0, 5.0$ & $\ge 0.0319$ & all diverged \\
\bottomrule
\end{tabular}
\end{table}

The boundary is sharp at $m \approx 0.5$--$0.8$. Choosing $\kappa = 0.25$
keeps Adam comfortably inside the stable region in all our experiments.

\section{Longer training schedule}
\label{sec:results-longrun}

The headline experiments use a deliberately short 10-epoch budget, where
ResNet-18 reaches about $81$--$84\%$ test accuracy with Adam (up to
$87\%$ for the strongest baseline), numbers that understate
what the network normally attains. To check that the short schedule does not
flatter (or undersell) the divergence-prevention result, we re-ran the full learning-rate grid for
100 epochs with cosine annealing, 3 seeds (Table~\ref{tab:longrun}).

\begin{table}[htbp]
\centering
\caption{Longer schedule on CIFAR-10/ResNet-18: 100 epochs with cosine
annealing, mean$\pm$std over 3 seeds. ``DIV.''\ = all seeds diverged
at the first update. At the realistic length the network reaches $\approx 93\%$;
Adam-InitOnly retains that accuracy across the full $10^{-3}$--$3.0$ range of
initial learning rates, while vanilla Adam still diverges for any
$\eta \ge 10^{-1}$.}
\label{tab:longrun}
\resizebox{\textwidth}{!}{%
\begin{tabular}{lcccccc}
\toprule
$\eta_{\text{init}}$ & $10^{-3}$ & $10^{-2}$ & $10^{-1}$ & $0.3$ & $1.0$ & $3.0$ \\
\midrule
Vanilla Adam  & $0.935 \pm 0.002$ & $0.931 \pm 0.001$ & DIV.\ & DIV.\ & DIV.\ & DIV.\ \\
Adam-InitOnly & $0.935 \pm 0.001$ & $0.930 \pm 0.004$ & $0.926 \pm 0.003$ & $0.926 \pm 0.001$ & $0.927 \pm 0.002$ & $0.925 \pm 0.002$ \\
\bottomrule
\end{tabular}%
}
\end{table}

At the optimal $\eta = 10^{-3}$ vanilla Adam and Adam-InitOnly are identical
($0.935$ both), confirming that the probe has no effect on training when the learning rate is
already well chosen. Across the entire range up to a $3000\times$
misspecified $\eta = 3.0$ (where vanilla Adam diverges in the first step and has no
entry to report for any $\eta \ge 10^{-1}$) Adam-InitOnly stays in
$[0.925, 0.935]$, within one percentage point of the optimal configuration at
every learning rate. The divergence prevention is therefore not an artefact of the short
10-epoch budget: at a realistic training length the protected model reaches
the same $\approx 93\%$ accuracy regardless of how badly the initial learning
rate was chosen.

\section{Per-architecture results for the direction-matched probe}
\label{app:kappa-perarch}

The aggregate rows of Table~\ref{tab:universal-kappa} average test accuracy over
nine architectures with different chance levels and task difficulties, so the
aggregate is meant only as a compact summary. Table~\ref{tab:kappa-perarch}
breaks it down per architecture at the discriminating $100\times$-too-large
rate, for the recommended $\kappa = 2$; Figure~\ref{fig:kappa-perarch} in
the main text visualises this table. At this setting the direction-matched
probe is the best variant on every architecture \emph{and never diverges},
including the plain MLP, the lone partial failure at $\kappa = 3$
(Table~\ref{tab:universal-kappa}), which is fully protected here.

\begin{table}[htbp]
\centering
\caption{Per-architecture test accuracy at the $100\times$-too-large learning
rate ($\eta = 0.1$), mean over 3 seeds; a divergent run counts as $0$,
``DIV.''\ = all seeds diverged. The Adam-direction probe uses the recommended
$\kappa = 2$ (vanilla Adam and the grad probe as in
Table~\ref{tab:universal-kappa}); it diverges on none of the nine
architectures. Suite budget: 3 epochs (Table~\ref{tab:protocol}) --
absolute accuracies are therefore well below the 10-epoch grids of
Table~\ref{tab:rescue} and not comparable to them.}
\label{tab:kappa-perarch}
\resizebox{\textwidth}{!}{%
\begin{tabular}{lccc}
\toprule
Architecture & vanilla Adam & grad probe & Adam-dir probe ($\kappa = 2$) \\
\midrule
ResNet-18 BN (CIFAR-10) & DIV. & $0.56$ & $\mathbf{0.68}$ \\
VGG BN (CIFAR-10) & 0.28\,(1/3 d.) & $0.65$ & $\mathbf{0.68}$ \\
ResNet-18 GN (CIFAR-10) & DIV. & $0.47$ & $\mathbf{0.53}$ \\
ResNet-18 no-norm (CIFAR-10) & DIV. & DIV. & $\mathbf{0.54}$ \\
CNN (Fashion-MNIST) & DIV. & DIV. & $\mathbf{0.90}$ \\
MLP (Fashion-MNIST) & DIV. & DIV. & $\mathbf{0.86}$ \\
LSTM (AG News) & 0.22\,(1/3 d.) & 0.08\,(2/3 d.) & $\mathbf{0.83}$ \\
Transformer no-warmup (AG News) & DIV. & $0.25$ & $\mathbf{0.82}$ \\
Transformer wide (AG News) & DIV. & $0.25$ & $\mathbf{0.83}$ \\
\bottomrule
\end{tabular}}
\end{table}

\section{Imagenette: a real-resolution control}
\label{app:imagenette}

This appendix gives the exact numbers behind the top-right panel of
Figure~\ref{fig:rescue-headline}.
Imagenette~\citep{howard2019imagenette} is the fastai ImageNet subset
(10 classes, 9{,}469/3{,}925 train/val images); we train the standard,
unmodified torchvision ResNet-18 (7$\times$7 stride-2 stem with
max-pooling) at $128 \times 128$ for 20 epochs (the small dataset
yields only 37 steps per epoch), all other settings as in
Section~\ref{sec:setup}. At the well-tuned rate the probe makes no
material difference ($0.767$ vs $0.769$, within the seed spread).

\begin{table}[htbp]
\centering
\caption{Imagenette at $128 \times 128$ with a standard ResNet-18
(20 epochs, mean$\pm$std over 5 seeds; best per column within each
block in bold). ``DIV.''\ = all seeds diverged at the first update. The
direction-matched probe uses the same fixed $\kappa = 2$ as everywhere
else.}
\label{tab:imagenette}
\resizebox{\textwidth}{!}{%
\begin{tabular}{lcccccc}
\toprule
$\eta_{\text{init}}$ & $10^{-3}$ & $10^{-2}$ & $10^{-1}$ & $0.3$ & $1.0$ & $3.0$ \\
\midrule
Vanilla Adam & $\mathbf{0.769 \pm 0.021}$ & $\mathbf{0.724 \pm 0.022}$ & DIV.\ & DIV.\ & DIV.\ & DIV.\ \\
\midrule
\multicolumn{7}{l}{\textit{Direction-matched probe, fixed $\kappa = 2$ (no calibration)}} \\
\textbf{Adam-InitOnly (dir.)} & $\mathbf{0.767 \pm 0.020}$ & $\mathbf{0.750 \pm 0.011}$ & $\mathbf{0.738 \pm 0.010}$ & $\mathbf{0.742 \pm 0.017}$ & $\mathbf{0.729 \pm 0.035}$ & $\mathbf{0.718 \pm 0.040}$ \\
\bottomrule
\end{tabular}}
\end{table}

\section{Experimental protocol}
\label{app:protocol}
Each result is tied to an integer seed. In the learning-rate sweep and
language-model experiments every run is launched as its own process,
and the seed controls both the data order and the model initialisation.
In the cross-architecture suite and the held-out check the data loaders
are built once with a fixed shuffle seed and shared across the runs of
the script; there the seed controls the model initialisation, while the
batch order varies from run to run. The per-cell results of these
tables therefore average over model seeds and independent batch orders,
and no batch-paired comparison across variants is claimed. The mean accuracy is based
on the number of seeds as specified in the captions: five seeds in the CIFAR-10
sweep experiment, three seeds in the cross-architecture experiment and
correlation experiment, four seeds in the default-unsafe divergence experiment,
and up to one hundred in the XOR control. The figures for wall-clock time overhead
are obtained through a single interleaved and GPU-warmed benchmark instead
of runs collected over sessions, which we found to be contaminated by external load.
The correlation experiment reports the line-search step against a
PyHessian/Lanczos estimation of $\lambda_1$ based on the same probe batch.
The reference implementation behind all experiments, including the
per-run probe outcomes of the suite, held-out and language-model
experiments and the no-op test script, is publicly available at
\expandafter\url\expandafter{\coderepo}.

\section{The probe is a bit-exact no-op when the cap does not bind}
\label{app:noop}
That the non-binding cap does not affect training holds pathwise, not just in expectation,
and this fact is checked directly. The trials conducted by the probe check
the loss using parameters that have been perturbed; the parameters
are always snapshotted and restored, but an implementation which only
restores the parameters leaks through two side-channels: the forward passes
in the trials update the running statistics for BatchNorm, and they consume
RNG state (dropout masks). The implemented probe snapshots and restores
both, and the restoration happens after the reading is done, so that the
returned $\bar{\alpha}$ is bit-identical to the unguarded implementation.

The verification uses pairs of seeds, an identical initial batch, and
deterministic kernels. For each of the nine suite architectures, we (i)
compute hashes of parameters, non-parameter
buffers, and the CPU+CUDA RNG state before and after a probe call, and (ii)
train one Adam step from the same seed both with and without a preceding
probe call, and compare resulting states bit-by-bit (running the test
without any probe serves as a determinism check). In all nine
architectures, the state hashes match after the probe call, and when the
cap does not bind, the first Adam update is bit-identical with and
without the probe call. Run on the pre-hardening version, the same
test procedure reveals the two leakage sources: BatchNorm buffers on the
two BN architectures and RNG state on the two dropout architectures,
together having a paired median test-accuracy effect of
$-0.001$, i.e.\ trajectory-level noise rather than a systematic advantage.
The test script and its per-architecture results ship with the reference
implementation.

The whole-run test is a more lenient one compared to the paired update,
as whole runs also depend on the determinism of the platform. At the single grid rate
where the cap does not bind ($\eta = 10^{-3}$ against
$\kappa\beta^{8} = 0.0078$), a ten-epoch AG News run with the hardened
probe perfectly mirrors the baseline run, seed by seed, whereas the
unguarded version that powers the reported numbers does not (median
$0.003$ in test accuracy). There are no non-deterministic kernels on that
benchmark, so that gap is the state leak. In the case of the
convolutional benchmarks no such claim can be made from either side:
re-running Adam against itself with the same random seed changes
CIFAR-10 test accuracy by a median of $0.008$ and by as much
as $0.018$. The differences between the probe and the baseline at that
particular rate are of the same magnitude irrespective of whether the
probe is hardened or not (medians $0.011$ and $0.006$ over the twenty
affected runs). Five seeds against that noise floor do not distinguish between the two
implementations. This is also why the rows where the directions match at
$\eta = 10^{-3}$ in Tables~\ref{tab:rescue} and~\ref{tab:imagenette} are
not exact copies of their baselines: the cap cannot bind there, so what is
left is just platform nondeterminism, together with the leak measured
above for the reported runs.

\end{document}